\documentclass[]{acmart}

\AtBeginDocument{%
  \providecommand\BibTeX{{%
    \normalfont B\kern-0.5em{\scshape i\kern-0.25em b}\kern-0.8em\TeX}}}

\setcopyright{none}

\usepackage{hyperref}
\usepackage{hyperxmp}

\usepackage{lscape}
\usepackage{multirow}
\usepackage{color}
\usepackage{fontawesome5}
\usepackage[dvipsnames]{xcolor}
\usepackage{todonotes}

\begin{document}

\title{A systematic review of relation extraction task since the emergence of Transformers}

\author{Célian Ringwald}
\email{celian.ringwald@inria.fr}
\orcid{1234-5678-9012}
\authornotemark[1]
\affiliation{%
  \institution{Université Côte d’Azur, Inria, CNRS, I3S}
  \city{Sophia-Antipolis}
  \country{France}
}

\author{Fabien Gandon}
\email{fabien.gandon@inria.fr}
\orcid{0000-0003-0543-1232}
\affiliation{%
  \institution{Université Côte d’Azur, Inria, CNRS, I3S}
  \city{Sophia-Antipolis}
  \country{France}
  }

\author{Catherine Faron}
\email{catherine.faron@inria.fr}
\orcid{0000-0001-5959-5561}
\affiliation{%
  \institution{Université Côte d’Azur, Inria, CNRS, I3S}
  \city{Sophia-Antipolis}
  \country{France}
}

\author{Franck Michel}
\affiliation{%
  \institution{Université Côte d’Azur, Inria, CNRS, I3S}
  \city{Sophia-Antipolis}
  \country{France}}

\author{Hanna Abi Akl}
\affiliation{%
  \institution{Data ScienceTech Institute, Inria}
  \city{Sophia-Antipolis}
  \country{France}
  }

\renewcommand{\shortauthors}{Ringwald and Gandon, et al.}

\begin{abstract}
This article presents a systematic review of relation extraction (RE) research since the advent of Transformer-based models. Using an automated framework to collect and annotate publications, we analyze 34 surveys, 64 datasets, and 104 models published between 2019 and 2024. The review highlights methodological advances, benchmark resources, and the integration of semantic web technologies. By consolidating results across multiple dimensions, the study identifies current trends, limitations, and open challenges, offering researchers and practitioners a comprehensive reference for understanding the evolution and future directions of RE.
 \begin{center}
   \line(1,0){70}
 \end{center}
The resulting annotated \href{https://www.zotero.org/groups/6070963/scilex_re_systlitreview/library}{\faGraduationCap
 Zotero bibliography} has been made openly available and reusable, together with the developed software \href{https://github.com/Wimmics/SciLEx}{\faGithub SciLEx}, which provides a unified framework for collecting and analysing papers in the context of systematic literature reviews.
\end{abstract}


\received{23 September 2025}

\maketitle

\section{Introduction}

The increasing digitalisation of daily life has created vast volumes of textual data that organisations and individuals must manage. Transforming this unstructured information into structured knowledge is a key challenge for natural language processing (NLP) and information systems. \textit{Relation Extraction} (RE) --- the task of identifying semantic relations between entities in a text --- addresses this challenge by producing triples of the form $(subject, relation, object)$. This structured representation enables more effective indexing, search, and retrieval, and provides the foundation for populating and augmenting knowledge bases (KBs) in general and knowledge graphs (KGs) in particular.

Knowledge graphs usage were popularised by Google as well as open-resources like Freebase~\cite{10.1145/1376616.1376746}, YAGO~\cite{10.1145/1242572.1242667}, DBpedia~\cite{10.1007/978-3-540-76298-0_52}, and Wikidata~\cite{10.1145/2629489}. They have become critical infrastructures in both academia and industry. They support applications ranging from search engines to recommendation systems and scientific discovery. However, maintaining and expanding these resources remains labour-intensive. Automatic relation extraction offers a promising mechanism for enriching KGs at scale, making RE both a foundational research task and a practical necessity.

\textit{Research Context and Objectives:} 
Research on RE spans more than three decades. A key milestone was the \textit{Fourth Message Understanding Conference} (MUC-4) in 1992, which introduced standard evaluation metrics (\textit{precision} and \textit{recall}) and the first RE system, \textit{Fastus}~\cite{hobbs-etal-1993-fastus}. Since then, RE has evolved from rule-based systems, to statistical learning, to neural architectures. The advent of Transformer-based pre-trained language models~\cite{devlin-etal-2019-bert,NEURIPS2020_1457c0d6} has marked a decisive turning point, delivering significant performance gains across a broad spectrum of NLP tasks~\cite{wang-etal-2018-glue} and revitalising research in knowledge acquisition and information extraction~\cite{DBLP:conf/emnlp/PetroniRRLBWM19,razniewski2021languagemodelsknowledgebases,pan_et_al:TGDK.1.1.2,zheng2024reliablellmsknowledgebases}.

\textit{Positioning with Respect to Previous Surveys:} 
Numerous surveys on RE have appeared in the past decade, but only a small subset adopt a rigorous systematic review methodology. Among them a great overview of the field was given by \citet{martinez-rodriguez_information_2020}, but it covers pre-2020 research and focuses on pipelines and rule-based methods. Subsequent works addressed narrower aspects, such as grammar-checking, where RE is a subtask~\cite{nasar_named_2021}, temporal relation extraction~\cite{yohan_bonescki_gumiel_temporal_2021}, or dataset creation and annotation practices~\cite{bassignana_what_2022}. More recent studies examine knowledge graph construction in general~\cite{schneider_decade_2022,lingfeng_zhong_comprehensive_2023}, but devote comparatively less attention to RE itself. These contributions remain valuable, but they are limited either in scope or methodology, leaving the field of RE without a comprehensive, systematic, and up-to-date synthesis.

\textit{Intended Audience:}  
This survey aims to give a concrete picture of the impact of Transformers on the Relation Extraction task, from researchers, students, but also any private researchers. It provides a reference for researchers and practitioners at the intersection of NLP and knowledge graphs, looking for a concrete understanding of the main research directions proposed to solve this task since the language models ramp-up. This survey also provides a first analysis that could be easily extended in specific directions since it is based on a public corpus composed of two levels: the complete set of papers collected, and a refined, annotated version thereof.

\textit{Survey purpose and Contributions:} 
This article presents an extended, systematic literature review (SLR) of RE from text, focusing on the period \textbf{2019--2024}, which corresponds to the rapid rise of Transformer-based models~\cite{10.5555/3295222.3295349}, and combining methodological rigour with comprehensive coverage. The main contributions of this review are: (1) An updated and detailed overview of RE research, highlighting the transformative impact of large pre-trained language models; (2) An application of the \textbf{SciLEx} framework, an extended SLR methodology ensuring transparency, reproducibility, and open availability of results; (3) A fine-grained annotation along more than thirty analytical dimensions across resources, models, and surveys, released openly to support future investigations.

\section{background}
\subsection{Knowledge Representation}


A knowledge graph (KG) is a graph linking entities (the nodes of the graph) with different types of binary predicates (the edges of the graph), also called relations or properties. Each edge of the graph links a subject entity to an object entity or literal value using a specific predicate. A graph can be viewed as a set of such triples (subject, predicate, object) that form the fundamental units of a knowledge.

Entities represent real-world objects or abstract concepts. Examples include people, locations, organisations, events, or ideas. When an entity can be uniquely identified, it is typically referred to as a \textit{named entity}.  Detecting entities from text generally relates to Named Entity Recognition (Section~\ref{def:ner}).
Attaching a unique identifier to an entity relates to the Entity Linking task (Section~\ref{def:EL}). 
Identifiers of entities are often IRIs (Internationalised Resource Identifiers)—an extension of URIs (Uniform Resource Identifiers) that supports Unicode.

A label is a human-readable string that serves as the descriptive name of an entity—e.g., “Paris”. Unlike identifiers, labels are not unique and may refer to multiple entities. Trying to identify labels of entities in a text is a task known as Mention Detection (Section~\ref{def:mention_detect}).

A relation connecting two entities is called an \textit{object property}; a relation connecting an entity to a literal value (such as a date, string, or number), is called a \textit{datatype property}. Identifying which relations are expressed by two entities in a text segment is generally part of the Relation Identification task (Section~\ref{def:RI}); attaching a unique identifier to a relation refers to the relation classification task (Section~\ref{def:RC}). Extracting triples from text is the special focus of our surveys and is generally referred as the Relation extraction task (Section~\ref{def:RE}).

Entities may also be associated with types, called classes or concepts. For instance Paris belongs the the class City or is of type City. Classes and properties are typically defined in an ontology, a schema, or more generally a vocabulary. Attaching a type to an entity is part of the NER task (Section~\ref{def:ner}), and is especially referred as the Entity Typing task (Section~\ref{def:ET}).

\subsection{Relation Extraction from Text}~\label{def:RE}

Let $D_b \subseteq W \times G$ be a dual dataset, where $W$ denotes a set of textual documents and $G$ represents the corresponding set of knowledge graphs. 
The relation extraction (RE) task can be formalized as learning an extraction function $E_{D_b}: W \rightarrow G; \; w \mapsto g$, where \( w \in W \) is an input text, and \( g \in G \) is the graph representing the relations and entities implied by \( w \).
It can be decomposed into the following subtasks:

\subsubsection{Mention Detection:}\label{def:mention_detect} This subtask consists in identifying specific spans of text within a document. Mention detection is a common task in natural language processing (NLP). In the context of RE, it includes recognising entity labels or extracting particular datatype properties. 
\textit{Example}: detecting the mentions \textit{"Victor Hugo"} and \textit{"Besançon"} in the text \textit{"Victor Hugo is born in Besançon"}.
    
\subsubsection{Entity Linking:}\label{def:EL} This subtask consists in assigning a unique identifier from a knowledge base (such as DBpedia or Wikidata) to a detected text span. 
\textit{Example}: \textit{"Paris is beautiful in spring."} contains the mention \textit{"Paris"} for which several entities are candidate (\(e_1 = \text{Paris, France}\), \(e_2 = \text{Paris, Texas}\), \(e_3 = \text{Paris Hilton}\)). The absence of "Hilton" in neighbouring tokens increases the likelihood of \(e_1\) leading  to the link \(\mathrm{EL}(\text{"Paris"}) = \texttt{http://dbpedia.org/resource/Paris}\).
    
\subsubsection{Coreference Resolution:}\label{def:CR} This subtask consists in identifying all expressions within a document that refer to the same entity and determining whether different mentions (such as names, pronouns, or noun phrases) are co-referential. 
\textit{Example}: \textit{"Marie Curie was a brilliant scientist. She discovered radium with her husband Pierre."} contains the mentions \(m_1 = \text{"Marie Curie"}\), \(m_2 = \text{"She"}\), and \(m_3 = \text{"her"}\), leading to a coreference cluster \(\{m_1, m_2, m_3\}\) referring to Marie Curie.

\subsubsection{Entity Typing:}\label{def:ET} This subtask consists in  assigning a semantic type to an entity. This type may be a class defined in an ontology or be drawn from a limited set of classes, e.g., $\{$Person, Place, Organisation$\}$.

\subsubsection{Named Entity Recognition (NER):}\label{def:ner} In the literature, NER often refers to mention detection combined with basic entity typing. However, some definitions also encompass entity linking and coreference resolution.
    
\subsubsection{Relation Identification:}\label{def:RI} Given a document and a tuple of previously identified entities, this subtask consists in determining whether a relation exists between the entities within the text. \textit{Example}: \textit{"Henri Matisse lived in Nice for many years."} contains the mentions \(m_1 = \text{"Henri Matisse"}\), \(m_2 = \text{"Nice"}\) and a relation identification output would include \(\mathrm{RI}(m_1, m_2) = \text{livedIn}\).

\subsubsection{Relation Classification:}\label{def:RC} This is a more advanced task than relation identification, where the specific type of relation expressed is identified, typically by assigning an identifier from a relation ontology or vocabulary. 
Following the previous example, a relation classification output could be \text{livedIn} $\equiv$ (\text{dbo:residence}, \text{wdt:P551}, \text{schema:homeLocation}), with relation identifiers from three well-known ontologies.\\

\noindent Other associated RE tasks include:

\subsubsection{Slot Filling:} Given a template with a missing value (a slot) and an input text, this task consists in extracting the correct value to complete the template. 
\textit{Example}: \texttt{Victor\_Hugo — bornIn — ?} applied to the text ``Barack Obama was born in Honolulu, Hawaii.'' leads to \(\text{bornIn}(\text{Barack Obama}, \text{Honolulu})\).

\subsubsection{Probing:} Popularised by the use of language models, this task involves completing an incomplete triple provided as input, often without specific context. Probing serves as a diagnostic task in NLP to evaluate what linguistic or factual knowledge is encoded in pretrained language models (e.g., BERT, GPT) by testing their performance on simple prediction tasks. 
\textit{Example}: \textit{What country is “Paris” in?} in the text \textit{“Paris is the capital of France.”} should lead to the output \textit{France}.

\subsubsection{Question Answering (QA):} Closely related to knowledge graph research, this task aims to produce an answer from a given question in natural language. This may involve generating queries, in SPARQL in the case of RDF knowledge graphs, which are subsequently executed against a knowledge graph to retrieve the answer.\\

Relation Extraction (RE) is therefore the task of identifying relationships between entities in a text, typically represented as triples of the form \((e_1, r, e_2)\).
In a \textbf{closed RE} setting, the task involves retrieving relations from a predefined set \(\mathcal{R}\), and optionally restricting entities to those referenced in a knowledge graph (KG) from a defined set \(\mathcal{E}\). This approach is well-suited for completing KGs but requires an ontology or, at least, a predefined set of relation types. It is typically implemented using supervised learning, which are unable to extract relations outside the fixed set \(\mathcal{R}\) (and optionally \(\mathcal{E}\)).
Conversely, in an \textbf{open RE} setting, the model is expected to produce triples using textual predicates, generally expressed as verbs or verb phrases extracted directly from the text: $\text{OpenRE}(s) \rightarrow \{(e_1, r_{\text{text}}, e_2)\}$.
This design is well-suited for bootstrapping and generating new knowledge from large corpora, and it is particularly useful in low-resource scenarios. However, the resulting data tends to be noisier and more redundant.
\section{SciLEx Methodology}~\label{sec:methodo}

The SciLEx\footnote{\url{https://github.com/Wimmics/SciLEx}} (Science Literature Exploration) review framework is a Python toolbox that implements and extends the methodology defined by~\cite{kitchenham2007guidelines,brereton_lessons_2007}. Starting from a user-defined keyword list, SciLEx automates the construction of a collection of relevant papers by generating and executing all possible combinations of queries derived from this keyword list across multiple digital libraries. This automation facilitates the paper collection process, ensures traceability, and supports the aggregation and deduplication of search results.
Additionally, SciLEx enriches the resulting corpus through the integration with external services such as PaperWithCode\footnote{available until may 2025 \url{https://paperswithcode.com} now redirects to Hugging Face}, CrossRef~\cite{hendricks_crossref_2020}, and Opencitation~\cite{peroni_opencitations_2020}. PaperWithCode, was intended for the Machine Learning community and aimed at connecting research articles to their corresponding methods, implemented code, evaluation results on standard datasets, and initial paper annotations. OpenCitation enables the retrieval of citations and references for a given paper, which can be used both to filter papers by impact and to expand the corpus through citation snowballing.
Finally, SciLEx exports all gathered information into a Zotero~\cite{mueen_ahmed_zotero_2011} collection[\href{https://www.zotero.org/groups/6070963/scilex_re_systlitreview/library}{\faClipboardList}], facilitating collaborative management, selection, and annotation of the corpus.

We began our work with an initial "tertiary study"~\cite{kitchenham_systematic_2010} (a review of reviews), followed by the systematic review of the models and datasets proposed to solve a RE task. 

\subsection{Digital Libraries Search Plan and Scope and Automated Metadata Enrichment}

We used SciLEx to collect articles on RE from nine digital libraries: HAL, Istex, arXiv, DBLP, Semantic Scholar, OpenAlex, Scopus, IEEE Xplore, and Springer Nature. Our search strategy relies on two keyword sets: 
$Kw_{RE} =$ \{"Relation extraction", "Relation Classification", "Triplet Extraction", "Slot Filling", "KG-to-Text", "Text-to-data extraction"\} and 
$Kw_{Survey} = $\{"Survey", "State-of-the-art", "Review", "Study"\}

For the tertiary study, we generated all possible combinations of RE keywords ($Kw_{RE}$) and survey-related terms ($Kw_{Survey}$). In the subsequent phase, the RE keywords alone ($Kw_{RE}$) were used to retrieve articles related to models and datasets without further distinction.
This comprehensive search strategy resulted in a total of 
$(6_{Kw_{RE}} \times 6_{Years} \times 9_{APIs}) + (6_{Kw_{RE}} \times 4_{Kw_{Survey}} \times 6_{Years} \times 9_{APIs}) = 2160$ queries executed via SciLEx. 
We deduplicated the retrieved articles by title and imported them into a Zotero library for further management. Ultimately, we collected more than 3,800 papers, including 3,567 proposed models, 250 datasets, and 77 survey articles.

\begin{figure}[ht!]
    \caption{Extended PRISMA flowchart: detailed process from the initial collection to the final selection and annotation}
    \label{fig:sample_figure}
\begin{minipage}[c]{\textwidth}
\centering
    \includegraphics[width=.6\linewidth]{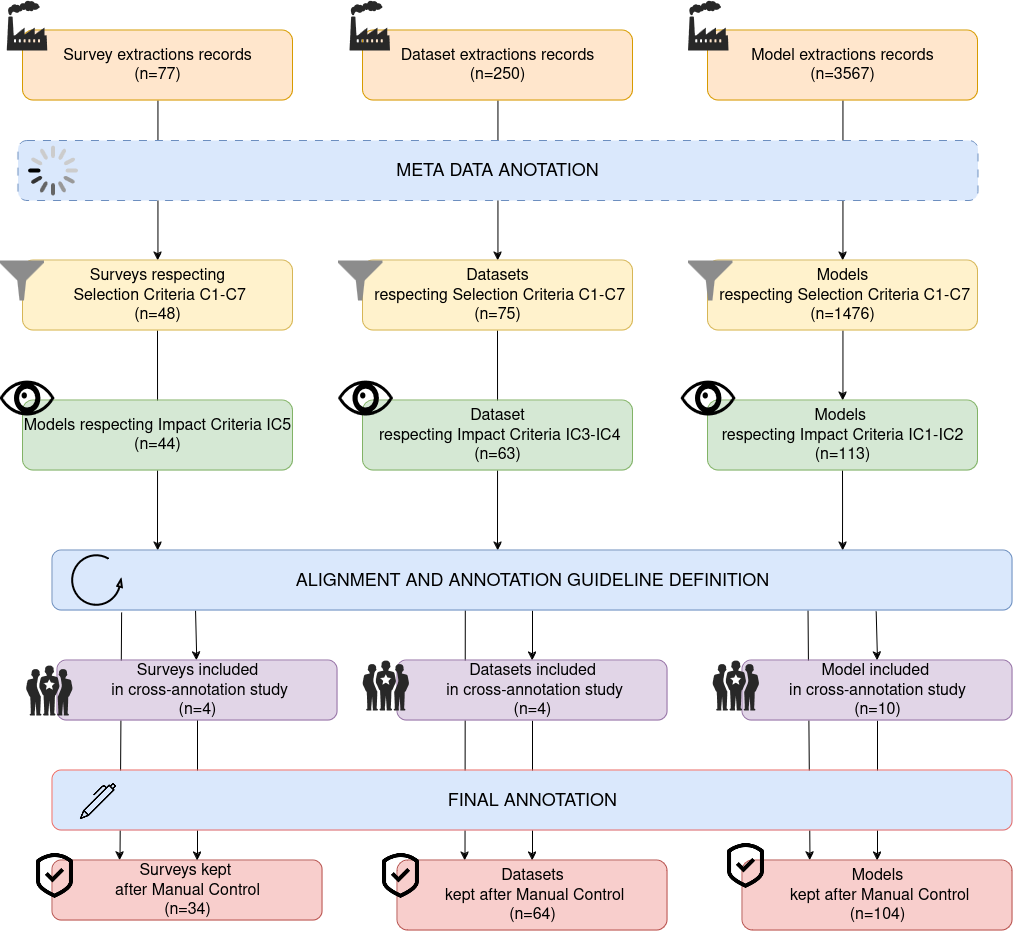}
\end{minipage}
\end{figure}
\sloppy
We enriched the resulting corpus with metadata from PaperWithCode\footnote{\url{https://paperswithcode.com}} and LinkedPaperWithCode\footnote{\url{https://linkedpaperswithcode.com}}, restricting inclusion to papers specifically related to the following RE sub-tasks:
$Kw_{PwC} =$ \{"Relation extraction", "Relation Classification", "Zero-shot Relation Triplet Extraction", "Zero-shot Relation Triplet Classification",
"Zero-shot Slot Filling", "Data-To-Text", "Text-To-Data"\}
From these sources, we obtained code repositories, benchmark results, and tags, which formed the initial basis for developing our annotation guidelines.
Subsequently, we retrieved the DOIs for all previously collected papers via CrossRef, and then gathered the citation network for each article using OpenCitations.

\subsection{Selection Criteria}\label{filter_and_annot}
We filtered the enriched corpus using inclusion criteria based on article metadata (see Tab.~\ref{tab:InclusionCrit}). We reduced the corpus to a consistent set of 1,476 models, 75 datasets and 48 surveys. 

\begin{table}[ht!]
\caption{Detailed inclusion criteria}
\label{tab:InclusionCrit}
\resizebox{\linewidth}{!}{
\begin{tabular}{|l|c|c|c|} \hline
Criteria & Exclusion criteria (EC) & Inclusion criteria (IC) &  Exceptions (Ex) \\ \hline
C1 - Publication period & \multicolumn{1}{l|}{Published between 2019 and 2024} & \multicolumn{1}{l|}{Not published between 2019 and 2024} & Inclusion: the paper is published before 2019 but is still relevant (often the case of dataset papers)\\ 
\hline
C2 - Coherence & \begin{tabular}[c]{@{}c@{}}Does not contain any \\survey term or task \\ related term \\ neither in the abstract \\ nor in the title\end{tabular} & 
\begin{tabular}[c]{@{}c@{}} Contains in the \\ abstract OR in the \textbf{title}\\ \textbf{at least}\\ \textbf{one} of the \textbf{survey terms}\\AND\\at least\\one of the \textbf{task-related terms}\end{tabular} 
& 
Inclusion: the paper does not contain any identified terms but is related to our RQs $\Rightarrow$  identification of new relevant terms \\
\hline
C3 - Language & in another language than English & the paper is written in English & / \\ 
\hline
C4 - Access & The access of the paper is not free & \begin{tabular}[c]{@{}c@{}} the paper could be read \\ without any paywall\end{tabular} & Inclusion:  somewhere accessible (partially open)\\ 
\hline
C5 - Publication process & the paper was not peer-reviewed & the paper was peer-reviewed & \begin{tabular}[c]{@{}c@{}}Inclusion: the paper was not peer-reviewed but very relevant~\\
Exclusion: the paper was peer-reviewed but not yet accessible \end{tabular} \\ 
\hline
C6 - Referenceability & the paper is not identified by a DOI & the paper has an associated DOI & Inclusion: the paper does not have a DOI but is very relevant \\ 
\hline
C7 - Duplicates & It exists a more recent version of this paper& \begin{tabular}[c]{@{}c@{}}The paper is the last version published\end{tabular} & if a paper is presenting both a model and a dataset it will be annotated twice \\
\hline
\multicolumn{1}{l}{} & \multicolumn{1}{l}{} & \multicolumn{1}{l}{} & \multicolumn{1}{l}{}
\end{tabular}
}
\end{table}


Because the corpus was still large, we applied additional impact-based selection criteria to retain only articles with significant influence in the scientific community:
\begin{itemize}
    \item IC1. Ranking of model Papers: 
    \begin{itemize}
        \item IC1.1 Journal publication: Impact factor greater than 10\footnote{We consolidated \href{https://www.scimagojr.com/}{SciMago} bibliometric data in our \href{https://github.com/datalogism/SciLex-RE_SystLitReview/blob/main/data/collect_filters/JournalImpactFactor.csv}{github}}
        \item IC1.2 Conference publication: ranked A* or A\footnote{According to \href{https://portal.core.edu.au/conf-ranks/}{CORE 2023}}
    \end{itemize}
    \item IC2. Citation of model papers: at least 1 citation per year since its publication
    \item IC3. Usage of dataset papers: used at least 2 times by one of the model papers studied.
     \item IC4. A dataset paper extending a dataset paper meeting criterion IC3 is accepted. 
     \item IC5. Survey papers impact: at least 3 citations.
\end{itemize}


\subsection{Annotation of the Collected Papers}

After selecting our corpus, we refined and extended the PaperWithCode tags to better address our research questions. The research team refined this process over four rounds, during which the annotation guidelines were discussed and iteratively improved.
%
A first round of annotation was conducted by one member of the team. 
This initial round provided a preliminary vocabulary of values for each dimension and served as a basis for discussions with the whole research team, whose members were later asked to cross-annotate a subset of each subcorpus. 
%
To evaluate the quality of our method and the consistency of the annotation dimensions proposed, we conducted a cross-annotation study on 10\% of the articles of each subcorpus:  we randomly selected a subset of 4 datasets, 4 surveys and 10 models among the articles already pre-annotated. This cross annotation validation implied 5 annotators, who each annotated 3 datasets, 2 surveys, and 6 models, in order to get 3 annotators per paper. 
We organized a three-hour meeting to launch the annotation process, where a short tutorial on the annotation platform and grid was presented to the annotators who could discuss during the process and were provided with Annotation Guidelines\footnote{\url{https://github.com/datalogism/SciLex-RE_SystLitReview/blob/main/doc/AnnotationGuidelinesV6.pdf}} including an annotation grid, a vocabulary of annotation dimensions, borderline examples and guidelines for solving ambiguous cases.

The annotation was prepared on a Zotero library, which enabled collaborative annotation and auto-completion. Each annotator was asked to spend at most 20 minutes per paper. They could add labels, update information, or delete labels as needed. Each expected annotation was indicated by a specific tag. All annotations were filled in as tags. They can be checked on the Zotero web interface by selecting a publication and opening the \textit{Tags} tab.

\begin{figure}[ht!]
    \caption{Number of annotation dimensions by Fleiss Kappa agreement level}
    \label{fig:agreement_distrib}
\begin{minipage}[c]{\textwidth}
\centering
    \includegraphics[width=.6\linewidth]{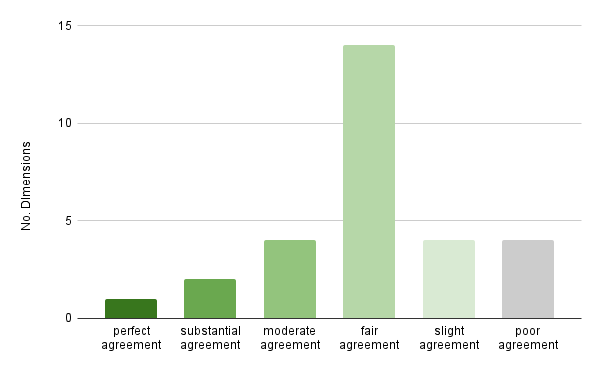}
\end{minipage}
\end{figure}

We considered the agreement of the annotation dimensions on the subset level (dataset/models/surveys). To our knowledge, no agreement metric considers the multi-annotators multi-possible answers scenario. The case of multiple possible answers was already studied on a one-by-one basis among inter-annotators. We propose in this work to consider each possible tag set given by the annotators and by annotation dimension as a possible answer to compute the Fleiss Kappa agreement on every annotation dimension. The Fleiss Kappa score is close to 1 when annotators fully agree, and a value close to -1 indicates a strong disagreement. The detailed results\footnote{\href{https://raw.githubusercontent.com/datalogism/SciLex-RE_SystLitReview/refs/heads/main/data/annotation_results/Fleiss_by_dim_type.csv}{Fleiss Kappa results available on github}} show that annotators are closer to agreement on the data related to the dataset, followed by survey data and finally the model one. We can note a perfect agreement on the type of survey (Survey methodology), and substantial agreement on some other dimensions among which the Pretrained model used (PTM) and  the Number of types of relation represented by datasets or targeted by the models. Other dimensions benefit from a moderate agreement, like the Use of negative examples, the Granularity, the Task and the Input type. Overall, as shown in Fig.~\ref{fig:agreement_distrib}, our annotations have a good level of agreement.   

A resolution stage allowed us to manually align and aggregate conflicts from the annotations. Conflicts were discussed in regular meetings until full agreement was reached, and the guidelines were adjusted to clarify how to handle complex or ambiguous cases.



The guidelines were revised in the light of all the issues reported during the cross-annotation study. On this basis a unique expert annotator was asked to complete the annotation on the full corpus. This step required significant effort, as annotating a paper took about 10 minutes on average, totaling more than 37 hours of work.

\begin{landscape}
\begin{table}[p]
\caption{Detailed Annotation Grid}\label{tab:AnnotGrid}
\resizebox{\linewidth}{!}{%
\begin{tabular}{|l|ll|lll|}
\hline
Dimension               & Desc.                                                                                                                                                                  & Datatype & Surveys    & Datasets   & Models     \\ 
\hline
Benchmark type          & Qualify the type of survey: qualitative or quantitative                                                                                                                   & Quali    & \checkmark &            &            \\
Survey Methodo          & Is the survey following a systematic review process?                                                                                                                  & Bin      & \checkmark &            &            \\
SemanticWebTechno       & Is the survey introducing the semantic web technologies in the context of RE?                                                                                         & Bin      & \checkmark &            &            \\ 
Lang                    & Which language is targeting the paper?                                                                                                                                & Quali    & \checkmark & \checkmark & \checkmark \\
Source                  & From which source is derived the paper?                                                                                                                               & Quali    & \checkmark & \checkmark &            \\
Domain                  & Is the article focused on a limited domain? (Finance, news, encyclopedic...)                                                                                          & Quali    & \checkmark & \checkmark & \checkmark \\
ManualAnnotation        & Did the conducted research involve manual annotation?                                                                                                                & Bin      & \checkmark & \checkmark & \checkmark \\
UseNER                  & Is the proposed process integrating tools to solve NER?                                                                                                               & Quali    &            & \checkmark &            \\
SynthGeneration\_Bin    & Is the paper using/creating synthetic data?                                                                                                                           & Bin      &            & \checkmark & \checkmark \\ 
NbTypeRel               & How many types of relation are covered?                                                                                                                                & Quanti   &            & \checkmark & \checkmark \\
NbEntity                & How many entities are covered?                                                                                                                                        & Quanti   &            & \checkmark & \checkmark \\
NbTypeEntity            & How many classes/types of entities are covered?                                                                                                                         & Quanti   &            & \checkmark & \checkmark \\
NbTriples               & How many triples in total are considered in the paper?                                                                                                                & Quanti   &            & \checkmark &            \\
NbDoc                   & How many documents are considered in the corpus? (if granularity=doc)                                                                                                 & Quanti   &            & \checkmark &            \\
NbSent                  & How many sentences are considered in the corpus? (if granularity=sent)                                                                                               & Quanti   &            & \checkmark &            \\
Granularity             & \begin{tabular}[c]{@{}l@{}}Are the discussed, proposed or used resources \\ given at the document level or paragraph/sentence level?\end{tabular}            & Quali    & \checkmark & \checkmark & \checkmark \\ 
Task                    & Which task is considered by the resource?                                                                                                                             & Quali    & \checkmark & \checkmark & \checkmark \\
Dataset                 & Which datasets are used in the paper?                                                                                                                & Quali    & \checkmark &            & \checkmark \\
DatatypeProp            & \begin{tabular}[c]{@{}l@{}}Which datatype are explicitly used in the object of triples? (String/Values/Dates...)\end{tabular}                                    & Quali    &            & \checkmark & \checkmark \\
DatasetSplit            & How was the dataset splitted? (Randomly, TimeAware...)                                                                                                                & Quali    &            & \checkmark & \checkmark \\
Input                   & Which content is taken in the proposed model as input? (Text/Graph/Images...)                                                                                            & Quali    &            &            & \checkmark \\
ObjectProperties\_Bin   & Are the objects of triples URIs?                                                                                                   & Bin      &            & \checkmark & \checkmark \\
UseNegativeExample\_Bin & Are the resources containing or using negative examples?                                                                                                              & Bin      &            & \checkmark & \checkmark \\
NbDataset               & Number of datasets considered in the paper                                                                                                                            & Quanti   & \checkmark & \checkmark & \checkmark \\
LinearizedGraph\_Bin    & Is the model explicitly defining a linearisation to represent graph as text?                                                                                          & Bin      &            &            & \checkmark \\ 
NbModel                 & Number of models considered in the paper                                                                                                                             & Quanti   & \checkmark &            & \checkmark \\
Archi                   & \begin{tabular}[c]{@{}l@{}}Which architectural layer is considered in the proposed work?\\ (Convolutional Network, MultiLayer Percepton, Transformer...)\end{tabular} & Quali    & \checkmark &            & \checkmark \\
PTM                     & Which pretrained model is used/described in the paper?                                                                                                               & Quali    & \checkmark & \checkmark & \checkmark \\
LearningMethod          & Which method is used to train the proposed model?                                                                                                                     & Quali    &            &            & \checkmark \\
DecodingMethod\_Bin     & Is the model explicitly updating the method used to generate the output?                                                                                             & Bin      &            &            & \checkmark \\
CostEval\_Bin           & Is the model evaluating the cost of the training process?                                                                                                                                & Bin      &            &            & \checkmark \\
LossUpdate\_Bin         & Is the proposed model using a custom Loss function during the training?                                                                                                        & Bin      &            &            & \checkmark\\
\hline
\end{tabular}}
\end{table}
\end{landscape}


In the end, we selected 32 annotation dimensions to describe our collected corpus of surveys, datasets, and models, among which 5 of them are derived from PaperWithCode. They are presented in Table~\ref{tab:AnnotGrid}. They may take three types of values: boolean (True/False), quantitative (expressed in powers of 10, e.g., $10^0 = [1-10]$, $10^1  = [10-100]$), and qualitative (open-ended values identified during the first annotation round). These dimensions are detailed in the Annotation Guideline. 

\section{Data Analysis of the Collected Papers}\label{sec:collection}

We collected 3,894 papers, gathering 77 surveys, 250 datasets papers and 3,567 model papers. We first annotated them with all the available basic bibliographic data. Applying selection criteria C1–C7 then reduced the corpus to 1,500 papers. Although this subset still included many model papers, their number was drastically reduced by applying the Impact Criteria. Annotation and manual verification ultimately narrowed the scope to 202 papers, among which 34 surveys, 64 datasets, and 104 models.

\begin{figure}[ht!]
\centering
\begin{minipage}{.45\textwidth}
\centering
\caption{Nb. of surveys published by year}
\label{fig:survey_pubyear}
\includegraphics[width=\linewidth]{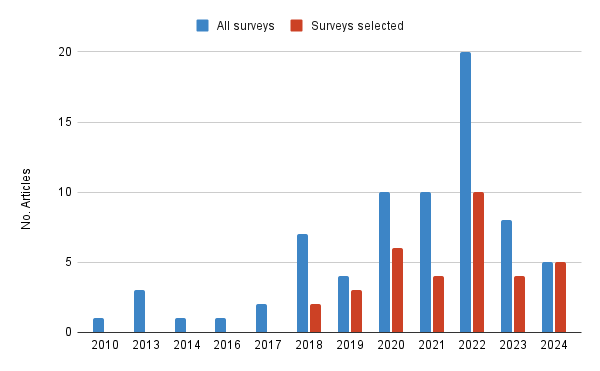}
\end{minipage}\hfill
\begin{minipage}{.45\textwidth}
\centering
\caption{Publication types of surveys}
\label{fig:survey_pubtype}
\includegraphics[width=\linewidth]{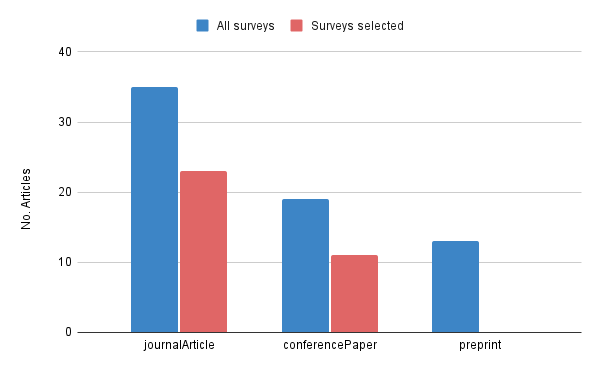}
\end{minipage}
\end{figure}

\begin{figure}[ht!]
\centering
\begin{minipage}{.45\textwidth}
\centering
\caption{Referencing of surveys}
\label{fig:survey_referencing}
\includegraphics[width=\linewidth]{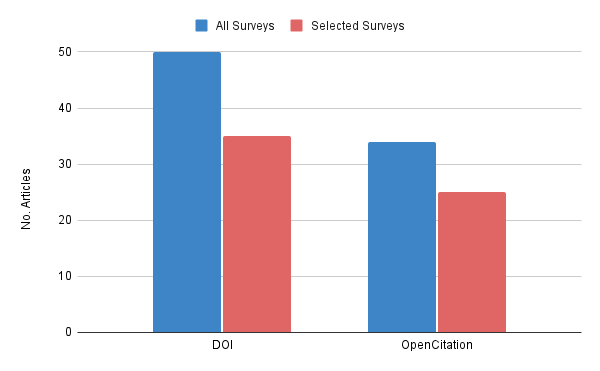}
\end{minipage}\hfill
\begin{minipage}{.45\textwidth}
\centering
\caption{Accessibility of surveys}
\label{fig:surveys_access}
\includegraphics[width=\linewidth]{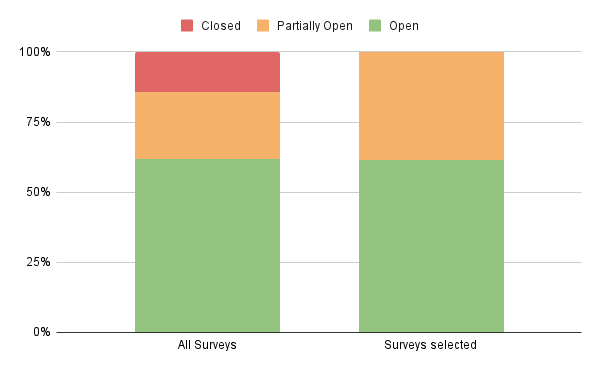}
\end{minipage}
\end{figure}
\textit{Surveys}: the 67 collected surveys were published between 2010 and 2024. Fig.~\ref{fig:survey_pubyear} shows that publications increased significantly after 2020, reaching 10 papers per year and the double in 2022. The subset analyzed in this survey is more evenly distributed between 2018 and 2024 and follows the same trend. As shown in Fig.~\ref{fig:survey_pubtype}, most collected surveys were published in journals, with about half as many appearing in conferences. Preprints formed a minor portion of the collected surveys and were excluded by our selection criteria. The selected surveys came from a diverse set of journals. ACM Computing Surveys is well represented as the journal is specialised in publishing systematic reviews. There are also journals specialised in Database (Data Technologies and Applications, Foundations and Trends in Databases), Semantic Web (Semantic Web Journal, Transactions on Graph Data and Knowledge) or general AI (IEEE Transaction on Neural Netorks, Learning Systems, IEEE Access). The main venues publishing surveys are computational linguistics conferences such as the Annual Meeting of the ACL and the conference on Empirical Methods in NLP (EMNLP). Fig.~\ref{fig:survey_referencing} shows that surveys are well-referenced, allowing the retrieval of a large part of the citation network associated to them. Please note that our selection process only kept surveys that were identified by a DOI. Finally,  Fig.~\ref{fig:surveys_access} shows that about half of the articles were either partially open or closed access.

\begin{figure}[ht!]
\centering
\begin{minipage}{.49\textwidth}
\centering
\caption{Nb. of dataset papers published by year}
\label{fig:dataset_pubyear}
\includegraphics[width=\linewidth]{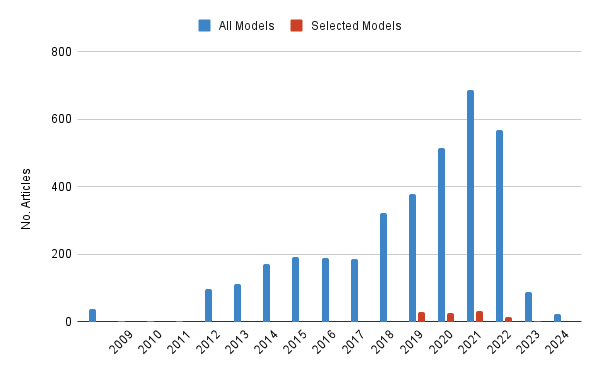}
\end{minipage}\hfill
\begin{minipage}{.49\textwidth}
\centering
\caption{Publication types of dataset papers}
\label{fig:dataset_type}
\includegraphics[width=\linewidth]{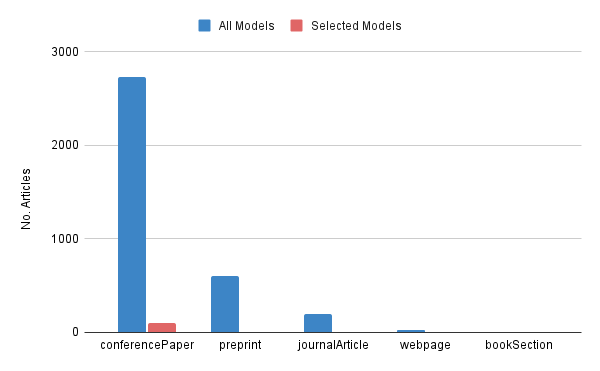}
\end{minipage}
\end{figure}

\begin{figure}[ht!]
\centering
\begin{minipage}{.45\textwidth}
\centering
\caption{Referencing of dataset papers}
\label{fig:dataset_ref}
\includegraphics[width=\linewidth]{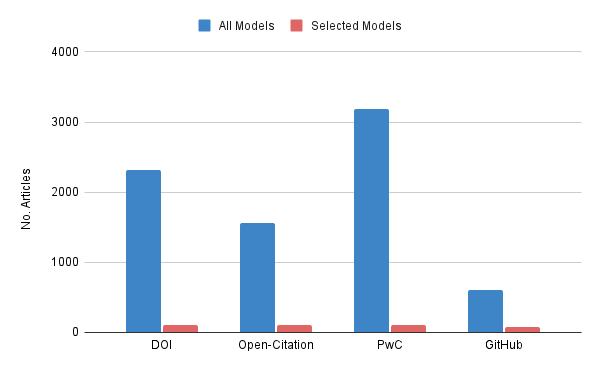}
\end{minipage}\hfill
\begin{minipage}{.45\textwidth}
\centering
\caption{Accessibility of dataset papers}
\label{fig:dataset_acces}
\includegraphics[width=\linewidth]{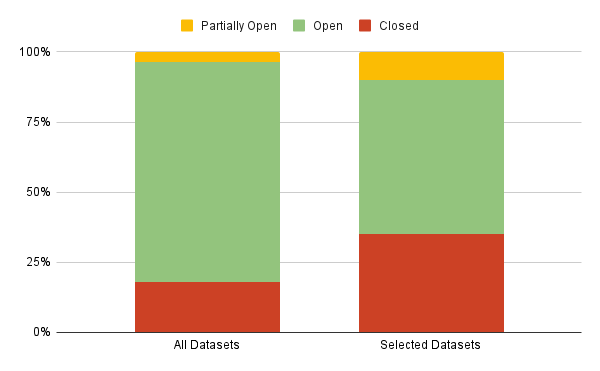}
\end{minipage}
\end{figure}

\textit{Datasets}: as shown in Fig.~\ref{fig:dataset_pubyear}, the 250 collected dataset papers span from 1990 to 2024. This wide temporal range reflects the enduring relevance of some benchmark datasets, which are still reused in later research. Dataset papers spiked around 2020, mirroring the increase in survey papers.
In contrast to surveys, dataset papers are predominantly introduced at conferences, especially through dedicated resource tracks (Fig.~\ref{fig:dataset_type}). The 64 dataset papers analysed in this systematic review are representative of both publication patterns and dataset characteristics. Several dataset papers are released outside traditional venues, often in workshops or working groups on specific research challenges. 
As shown in Fig.~\ref{fig:dataset_ref} most dataset papers have DOIs and many are indexed in OpenCitations. Approximately half of the dataset papers are listed on PaperWithCode, often linking to associated GitHub repositories. However, as shown in Fig.~\ref{fig:dataset_acces}, dataset access is inconsistent and may be restricted, similarly to surveys. 
Analysing the publication venues for the 64 dataset papers reviewed, we observe a substantial prevalence of computational linguistics conferences such as EMNLP and ACL. Additionally, workshops and shared tasks—most notably SemEval—play a significant role in dataset dissemination. When published in journals, datasets are usually related to biosciences and appear in journals venues such as Database (Oxford), Journal of Biomedical Informatics, or Briefings in Bioinformatics.

\begin{figure}[ht!]
    \centering
\begin{minipage}{.45\textwidth}
\centering
\caption{Nb. of model papers published by year}
\label{fig:models_pubyear}
\includegraphics[width=\linewidth]{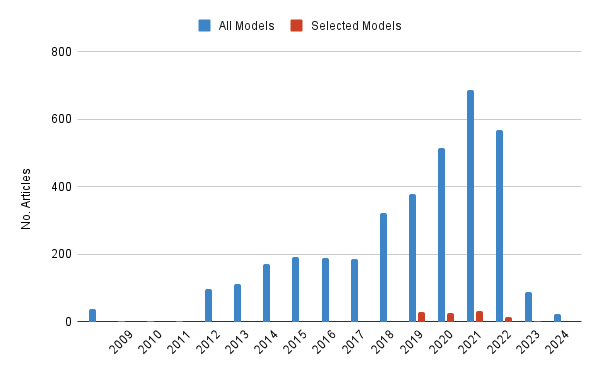}
\end{minipage}\hfill
\begin{minipage}{.45\textwidth}
\centering
\caption{Publication types of model papers}
\label{fig:models_types}
\includegraphics[width=\linewidth]{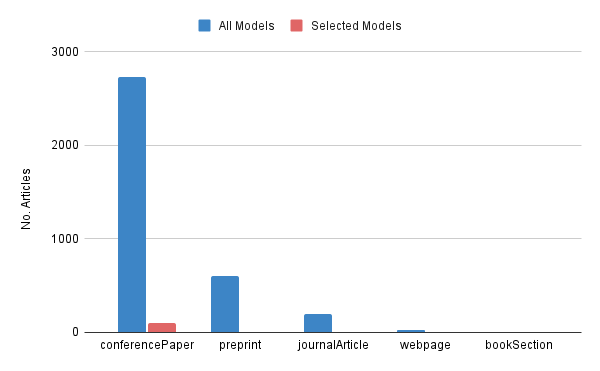}
\end{minipage}
\end{figure}
\begin{figure}[ht!]
\centering
\begin{minipage}{.45\textwidth}
\centering
\caption{Referencing of model papers}
\label{fig:models_ref}
\includegraphics[width=\linewidth]{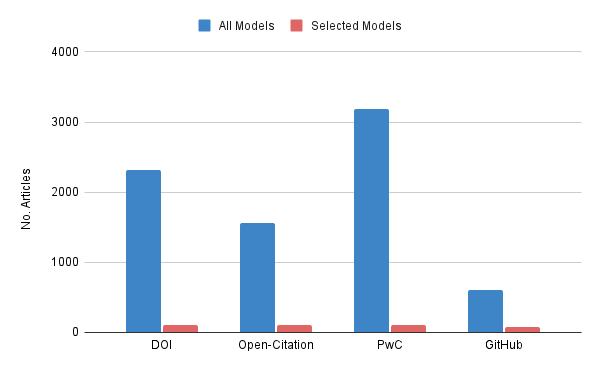}
\end{minipage}\hfill
\begin{minipage}{.45\textwidth}
\centering
\caption{Accessibility of model papers}
\label{fig:models_acces}
\includegraphics[width=\linewidth]{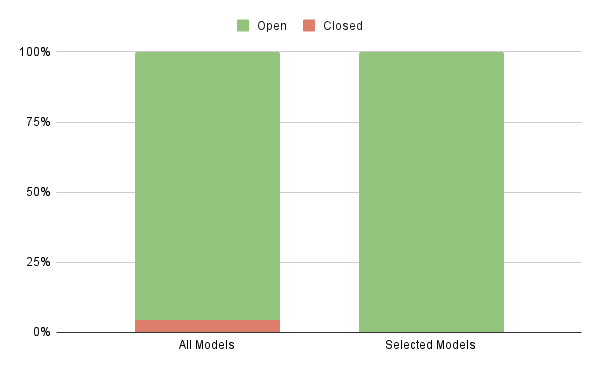}
\end{minipage}
\end{figure}

\textit{Models}: as shown in Fig.~\ref{fig:models_pubyear}, the 3,567 model papers collected were published between 2009 and 2024. A clear increase in RE publications is observed: about 200 papers were published between 2014–2017, rising to 400 in 2019, and peaking at over 600 in 2021.
Fig.~\ref{fig:models_types} shows that models are rarely published in journals; most appear in conferences such as ACL, EMNLP, IJCAI, or AAAI. Regarding their indexing and referencing (see Fig.~\ref{fig:models_ref}), not all models are associated with a DOI, and citation network data could only be retrieved for approximately half of the papers. 
According to PaperWithCode metadata, all collected models are listed, but only a minority provide access to source code or benchmark datasets. In terms of accessibility, most articles are openly available, as shown in Fig.~\ref{fig:models_acces}.

\section{Analysis of Surveys on Relation Extraction}
Our analysis relies on the curated collection of 34 surveys on the relation extraction (RE) task published between 2019 and 2023. This represents 44\% of the corpus of surveys initially collected.
Surveys on RE vary considerably in scope, methodology and depth. 
Out of the 34 surveys in our curated corpus, 12 are purely qualitative.
These are typically concise papers, often published on \textit{arXiv} and covering a limited number of datasets (less than ten) while discussing between 10 and 100 models. Their primary purpose is to provide broad overviews of research activity rather than detailed methodological or benchmarking analyses. By contrast, thirteen surveys attempt to combine performance results with descriptive analysis of models, while five focus exclusively on benchmarking. From a methodological perspective, only seven surveys (20\% [\href{https://www.zotero.org/groups/6070963/scilex_re_systlitreview/collections/C3UQ4BLW/tags/SURVEY_METHODO_BIN%3A1/collection}{\faClipboardList}]) apply principles consistent with a Systematic Literature Review (SLR), representing a minority of the corpus. These are directly compared with our work in Table~\ref{tab:compare_surveys}.
A few studies on RE have been published after our corpus was finalized, among which~\cite{10.1145/3674501} that converges on some aspects with our survey, but is more limited in terms of paper annotation and does not rely on a detailed methodology.
 

\begin{table}[ht!]
\centering
\caption{Comparison of our work with previous systematic literature reviews. To gain in readability, we refer to tasks using abbreviations: NER, for Named Entity Recognition, CR: Coreference Resolution, ET: Entity Typing, EL: Entity Linking, RC: Relation Classification. Also, Sent and Doc refer to sentence and of document granularity, respectively. "?" refers to the case where the information is not available or clearly stated in the survey. $\emptyset$ indicates that the survey is not concerned by the analysis dimension.}
\label{tab:compare_surveys}
\resizebox{\linewidth}{!}{
\begin{tabular}{|c|l|c|l|l|l|l|l|l|l|l|l|c|l|l|c|c|l|} 
\hline
\multicolumn{1}{|c|}{Year} & \multicolumn{1}{c|}{Survey} & Description & \multicolumn{7}{c|}{Task} & Architecture & PTM & Language & Granularity & Domain & Nb Models & Nb Datasets & Sem. Web \\ 
\cline{4-10}
 &  &  & NER & CR & ET & EL & RC & RE & Other &  &  &  &  &  &  &  &  \\ 
\hline
2020 & \cite{martinez-rodriguez_information_2020} & Complete overview of research before 2020 & \checkmark & \checkmark & \checkmark & \checkmark & \checkmark & \checkmark &  & Pipeline, RuleBased & $\emptyset$ & En & Doc+Sent & Multi & $10^1$ & $10^1$ & \checkmark \\ 
2021 & \cite{nasar_named_2021} & RE in grammar checking & \checkmark & \checkmark & \checkmark & \checkmark &  & \checkmark &  & Transformer, SVM & Elmo & Multi & Document &  & $10^1$ & $10^1$ &  \\ 
2021 & \cite{yohan_bonescki_gumiel_temporal_2021} & Temporal relation extraction &  &  &  &  & \checkmark & \checkmark & \checkmark & NeuralNet, RandomForest & Bert & EN,FR & Doc+Sent & Medical & $10^2$ & $10^1$ &  \\ 
2022 & \cite{bassignana_what_2022} & Dataset creation and annotation & \checkmark &  &  &  & \checkmark & \checkmark &  & NeuralNet & Bert, Glove, SciBert & EN+Multi & Doc+Sent & Multi & $10^1$ & $10^1$ &  \\ 
2022 & \cite{schneider_decade_2022} & Global overview of research output & \checkmark &  &  & \checkmark &  &  & \checkmark & Knowledge Augmented LM & ERNIE, K-BERT, KEPLER, COMET & ? & $\emptyset$ & Multi & ? & 0 &  \\ 
2023 & \cite{lingfeng_zhong_comprehensive_2023} & Horizontal overview of KG+NLP & \checkmark & \checkmark & \checkmark & \checkmark & \checkmark & \checkmark &  & Transformer, NeuralNet & BERT, Elmo, LUKE & Multi & Doc+Sent & Multi & $10^1$ & $10^1$ &  \\ 
2025 & Our work & Horizontal + vertical overview of KG+NLP & \checkmark & \checkmark & \checkmark & \checkmark & \checkmark & \checkmark & \checkmark & NeuralNet, Transformer & Many & Multi & Doc.+Sent & Multi & $10^2$ & $10^1$ & \checkmark \\
\hline
\end{tabular}
}
\end{table}

The Systematic reviews in our corpus describe core RE subtasks (e.g., \textit{NER}, \textit{Coreference Resolution}, \textit{Entity Linking}, \textit{Relation Classification}), as well as dimensions such as language, granularity and application domain. However, none of them provide direct access to their annotated corpora; such information must generally be inferred from the text. By contrast, our work openly releases an extended analytical grid with more than thirty dimensions, enabling fine-grained analysis and reproductibility. In addition, most surveys focus on classical neural architectures with embeddings, which are increasingly outdated in light of the latest developments in transformer-based and graph neural models. The analysis presented in the surveys also tend to remain limited to small samples of roughly twenty models and datasets. Furthermore, they generally present datasets and models separately, without investigating their combined effect on the task formalisation. Our review addresses these gaps by jointly analysing models, datasets and their interactions, while also considering the role of Semantic Web resources, as initially proposed by \cite{martinez-rodriguez_information_2020}. Finally, our survey comes with a complete access to the collected bibliography\footnote{\url{https://github.com/datalogism/SciLex-RE_SystLitReview}}, as well as the tools used to conduct it\footnote{\url{https://github.com/Scilex}}, and a direct access to the data analysed\footnote{\url{https://www.zotero.org/groups/6070963/scilex_re_systlitreview/library}}.

Some surveys are linked to a benchmark. Eighteen surveys in our corpus report performance metrics, but only two offer substantial benchmarking depth. For example~\cite{Nayak_2021} enriches 4 datasets (NYT11-HRL, TACRED, WebNLG and SemEval-2010 Task 8) with results for 36 models, while \cite{ye-etal-2022-generative} extends the evaluation on the \textbf{NYT} dataset with 11 models. We build on these efforts by consolidating results from PapersWithCode (PWC), ensuring broader coverage and consistent comparisons across datasets and models. The detail of the results is available online\footnote{\url{https://github.com/datalogism/SciLex-RE_SystLitReview/blob/main/data/benchmark_data/Benchmark_consolidated.csv}}.

\section{Analysis of Papers presenting Datasets for the Relation Extraction Task}
Our analysis relies on the curated collection of 64 data papers published between 2019 and 2024, addressing the relation extraction (RE) task. This represents 25\% of the corpus of data papers initially collected. 

\textit{Granularity of Annotation:}
A key observation concerns the granularity at which relations are annotated. Half of the datasets (33 out of 64) are designed exclusively for paragraph-level RE,
while 11 target the document level. Here are the dataset papers dealing with document-level RE~[\href{https://www.zotero.org/groups/6070963/scilex_re_systlitreview/collections/JDPEMF65/tags/GRANULARITY%3ADocument/collection}{\faClipboardList}] and the those with sentence-level RE~[\href{https://www.zotero.org/groups/6070963/scilex_re_systlitreview/collections/JDPEMF65/tags/GRANULARITY%3ASentences/collection}{\faClipboardList}].
Restricting relation extraction to individual sentences limits the capture of context-long and interdependent sentential relations. To address this limitation, the remaining 20 datasets support both paragraph- and document-level annotations, allowing for a broader and more realistic representation of semantic relations in text.

\textit{Language Coverage:}
The vast majority of datasets (approximately 80\%) are dedicated to English~[\href{https://www.zotero.org/groups/6070963/scilex_re_systlitreview/collections/JDPEMF65/tags/LANG%3AEnglish/collection}{\faClipboardList}], indicating a clear research focus on monolingual RE in high-resource settings. Nevertheless, there has been a growing effort to expand linguistic coverage, with several datasets incorporating additional languages such as Arabic ~\cite{doddington_automatic_2004,walker_christopher_ace_2006,cabot_redrm_2023}, Chinese (in 5 papers, including~\cite{ellis_overview_2015,cheng_hacred_2021}), and Spanish (in 4 papers, including~\cite{uzzaman_semeval-2013_2013,noauthor_x-wikire_2019}) alongside English. This trend reflects an increasing awareness of the importance of multilingual information extraction.

\textit{Annotation Strategy:}
Given the need of deep learning models for large-scale training data, \textbf{distant supervision} has become a popular annotation strategy. While this approach enables rapid dataset construction, it often introduces noise in labels. To mitigate this, several initiatives have been proposed to annotate and/or correct datasets built using distant supervision.
In our corpus, 12 datasets were partially annotated to assess data quality~[\href{https://www.zotero.org/groups/6070963/scilex_re_systlitreview/collections/JDPEMF65/tags/MANUALANNOTATION%3APartial/collection}{\faClipboardList}], whereas 29 were entirely manually annotated~[\href{https://www.zotero.org/groups/6070963/scilex_re_systlitreview/collections/JDPEMF65/tags/MANUALANNOTATION%3A1/collection}{\faClipboardList}].

\textit{Source Material and Knowledge Base Integration:}
Wikipedia emerges as a dominant source of text, underpinning 23 datasets in our collection~[\href{https://www.zotero.org/groups/6070963/scilex_re_systlitreview/collections/JDPEMF65/tags/SOURCE%3AWikipedia/collection}{\faClipboardList}]. These are typically paired with structured knowledge bases such as DBpedia, YAGO, Wikidata, or Freebase, which provide the ground truth for relation labels. In earlier datasets, relations were commonly expressed as raw strings, particularly when constructed without knowledge base alignment. In the case of Wikidata, where entities are represented by opaque machine-readable identifiers (Q-ids), identifiers are generally replaced by human-readable labels to facilitate extraction and interpretation.

\textit{Types of Extracted Values:}
While string-based relations remain the most common in the RE datasets (54 datasets)~[\href{https://www.zotero.org/groups/6070963/scilex_re_systlitreview/collections/JDPEMF65/tags/DATATYPEPROP%3AString/item-list}{\faClipboardList}], there is notable attention to date-related relations (13 datasets)~[\href{https://www.zotero.org/groups/6070963/scilex_re_systlitreview/collections/JDPEMF65/tags/DATATYPEPROP%3ADate/item-list}{\faClipboardList}]. Only a minority (7 datasets)~[\href{https://www.zotero.org/groups/6070963/scilex_re_systlitreview/collections/JDPEMF65/tags/DATATYPEPROP%3AValues/item-list}{\faClipboardList}] focus on extracting numerical values. Aside it we can note that 28 datasets explicitly use object properties~[\href{https://www.zotero.org/groups/6070963/scilex_re_systlitreview/collections/JDPEMF65/tags/OBJECTPROPERTIES_BIN%3A1/item-list}{\faClipboardList}] accounting for only $1/3$ of the datasets proposed.

\textit{Domain:} 
Half of the datasets contain encyclopedic knowledge, and a quarter are news-related datasets. Scientific and biology-related datasets are also well represented in the corpus we analysed.
Some other popular datasets are related to specific contexts, such as conversational data (\textbf{ACE}) and cybersecurity (\textbf{CASIE}~\cite{satyapanich_casie_2020}). There are also datasets spanning multiple domains, among which \textbf{CROSS-RE}~\cite{noauthor_crossre_2022}, a fully annotated resource, and \textbf{MultiCROSSRE}~\cite{bassignana_multi-crossre_2023}, an augmented version created through translation.


\subsection{Encyclopedic Datasets}

\begin{table}[ht!]
\caption{Details on Encyclopedic datasets. To gain in readability, we use abbreviations to refer to tasks: NER: for Named Entity Recognition, CR: Coreference Resolution, ET: Entity Typing, EL: Entity Linking, RC: Relation Classification and GtoT: Graph-to-Text, as well as to some recurring sources: WP: Wikipedia, WD: Wikidata, DBP: DBPedia and FB: FreeBase. Also, Sent and Doc refer to sentence and of document granularity, respectively. "?" refers to cases where data is not available or clearly stated in the articles.}
\centering
\resizebox{\linewidth}{!}{
\begin{tabular}{|l|l|llllllll|l|l|l|ll|l|l|}
\hline
\multicolumn{1}{|l|}{Year} & Short Title                          & \multicolumn{8}{c|}{TASK}                                                                                                                                                                                           & \multicolumn{1}{c|}{GRANULARITY} & \multicolumn{1}{c|}{LANG} & \multicolumn{1}{c|}{SOURCE} & \multicolumn{2}{c|}{Datatype}               & \multicolumn{1}{l|}{ManualAnnot.} & \multicolumn{1}{c|}{UseNERTools} \\ \cline{3-10} \cline{14-15}
\multicolumn{1}{|l|}{}     &                                      & \multicolumn{1}{l|}{NER} & \multicolumn{1}{l|}{CR} & \multicolumn{1}{l|}{ET} & \multicolumn{1}{l|}{EL} & \multicolumn{1}{l|}{RC} & \multicolumn{1}{l|}{RE} & \multicolumn{1}{l|}{GtoT} & \multicolumn{1}{l|}{Other} & \multicolumn{1}{c|}{}            & \multicolumn{1}{c|}{}     & \multicolumn{1}{c|}{}       & \multicolumn{1}{l|}{ObjProp} & DtProp       & \multicolumn{1}{l|}{}             & \multicolumn{1}{c|}{}            \\ \hline
2011                       & WIKI-KBP\cite{riedel_modeling_2010} & \checkmark &                         &                         &                         &                         & \checkmark &                           &                            & Sent                             & EN                        & FB,NYT                      & \checkmark     & Str          &                                   & StandfordNER                     \\
2015                       & GoogleRE\cite{noauthor_GRE_2015}                             & \checkmark & \checkmark &                         & \checkmark &                         & \checkmark &                           &                            & Sent                             & EN                        & WP                          & \checkmark     & Str          & Partial                           &                                  \\
2017                       & WebNLG\cite{gardent_creating_2017}                               & \checkmark &                         & \checkmark & \checkmark &                         & \checkmark & \checkmark  &                            & Sent                             & EN                        & DBP                         & \checkmark     & Date,Str     & \checkmark                           &                                  \\
2017                       & WIKI-KB\cite{sorokin_context-aware_2017}                              & \checkmark &     
&                         & \checkmark &                         & \checkmark &                           &                            & Sent                             & EN                        & WD,WP                       & \checkmark                          & Date,Str     & Partial                           & Standfordnlp,Wikilinks           \\
2018                       & TREX\cite{elsahar_t-rex_2018}          & \checkmark & \checkmark &                         & \checkmark &                         & \checkmark &                           &                            & Sent                             & EN                        & WD,WP,DbP                   & \checkmark     & Str,Date,Val & Partial                           & Dbpediaspotlight                 \\
2018                       & FewRel\cite{han_fewrel_2018}                               & \checkmark & \checkmark &                         & \checkmark &                   \checkmark      & \checkmark &                           &                            & Sent                             & EN                        & WD,WP                       & \checkmark     & Str          & \checkmark          & Spacy, Wikilinks                            \\
2018                       & GDS\cite{jat_improving_2018}                                  &                          &                         &                         &                         &                         & \checkmark &                           &                            & Sent                             & EN                        & GOOGLERE                    &                              & Str          & \checkmark          &                                  \\
2019                       & DocRED\cite{yao_docred_2019}                               & \checkmark & \checkmark & \checkmark & \checkmark &                         & \checkmark &                           &                            & Doc                              & EN                        & WD,WP                       & \checkmark     & Date,Str,Val & Partial                           & Spacy                            \\
2019                       & FewRel 2.0\cite{gao_fewrel_2019}                               & \checkmark & \checkmark &  & \checkmark  &      \checkmark                   & \checkmark &                           &                            & Sent.                              & EN                        & WD,WP,FewRel, PubMed, UMLS                       & \checkmark    & Str. & \checkmark                           & ?                            \\
2019                       & X-WikiRE\cite{noauthor_x-wikire_2019}                            &                          &                         &                         &                         &                         & \checkmark &                           & \checkmark   & Sent                             & EN,FR,DE,ES               & WD,WP                       &                              & Str          &                                   &                                  \\
2019                       & LIFELONGFEWREL\cite{wang_sentence_2019}                       &                          &                         &                         &                         & \checkmark & \checkmark &                           & \checkmark   & Sent                             & EN                        & FewRel,SimpleQuestions      &                        & Str          &                                   &                                  \\
2019                       & WIKI-NRE/GEO-NRE\cite{trisedya_neural_2019}                     & \checkmark & \checkmark &                         & \checkmark & \checkmark & \checkmark &                           &                            & Sent                             & EN                        & WD,WP                       & \checkmark                           & Str          &                                   & Wikilinks                        \\
2019                       & KnowledgeNet\cite{mesquita_knowledgenet_2019}                         & \checkmark & \checkmark &                         & \checkmark &                         & \checkmark &                           &                            & Sent                             & EN                        & DIffBotKG,WD,WP             &                              & Date         & \checkmark          &                                  \\
2020                       & KILT\cite{noauthor_kilt_2020}                                 &                          &                         & \checkmark & \checkmark &                         & \checkmark &                           & \checkmark   & Doc+Sent                         & EN                        & Many                        &                              & ?            &                                   &                                  \\
2021                       & REBEL\cite{huguet_cabot_rebel_2021}                                &                          &                         &                         &                         &                         & \checkmark &                           &                            & Doc+Sent                         & EN                        & WD,WP                       & \checkmark     & Date,Str     &                                   & Wikilinks                        \\
2021                       & HacRED\cite{cheng_hacred_2021}                               & \checkmark &                         &                         & \checkmark & \checkmark &                         &                           &                            & Doc                              & CN,EN                     & CN-DbP,WP                   &                              & Str          & Partial                           & toolkit TexSmart                 \\
2021                       & WikiGraphs\cite{noauthor_wikigraphs_2021}                           & \checkmark &                         &                         & \checkmark &                         & \checkmark &                           &                            & Doc                              & EN                        & FB,WikiText-103             & \checkmark     & Str          &                                   &                                  \\
2021                       & CodRED\cite{noauthor_codred_2021}                               & \checkmark &                         &                         &                         &                         & \checkmark &                           &                            & Doc+Sent                         & EN                        & WD,WP                       & \checkmark     & None         & \checkmark          &                                  \\
2021                       & WIKI-EVENTS\cite{li_document-level_2021}                          & \checkmark &                         &                         &                         &                         &                         &                           & \checkmark   & Doc                              & EN                        & WP                          & ?                            & Str          & \checkmark          &                                  \\
2021                       & WIKI-ZSL\cite{chen_zs-bert_2021}                             & \checkmark &                         &                         &                         & \checkmark &                         &                           &                            & Sent                             & EN                        & Wiki-Kb,WD,WP               & \checkmark                           & Str          &                                   & Wikilinks                        \\
2022                       & ReDocRED\cite{tan_revisiting_2022}                             & \checkmark & \checkmark &                         &                         &                         & \checkmark &                           &                            & Doc                              & EN                        & Wiki-Kb,WD,WP               &                              & Str          &                                   &                                  \\
2022                       & HyperRED\cite{chia_dataset_2022}                             & \checkmark & \checkmark &                         & \checkmark &                         & \checkmark &                           &                            & Sent                             & EN                        & WD,WP                       & \checkmark     & Date,Val     & \checkmark          & DBPSpotlight,Spacy           \\
2022                       & MAVEN-ERE\cite{noauthor_maven-ere_2022}                            &                          & \checkmark &                         &                         &                         &                         &                           &                            & Doc                              & EN                        & WP                          & ?                            & Date,Str     & Partial                           &                                  \\
2022                       & WDV\cite{amaral_wdv_2022}                                  &                          &                         &                         &                         &                         & \checkmark & \checkmark  &                            & Sent                             & EN                        & WEBNLG,WP                   & \checkmark                          & Date,Str,Val & Partial                           &                                  \\
2023                       & DocRED-FE\cite{noauthor_docred-fe_2023}                            & \checkmark &                         &                         & \checkmark &                         & \checkmark &                           &                            & Doc                              & EN                        & WD,WP                       & \checkmark     &              & \checkmark          &                                  \\
2023                       & RELD\cite{pesquita_reld_2023}                                 & \checkmark &                         &                         & \checkmark & \checkmark & \checkmark &                           &                            & Doc+Sent                         & EN                        & Many                        &                              & Str          &                                   &                                  \\
2023                       & REDfm\cite{cabot_redrm_2023}                                &                          &                         & \checkmark &                         &                         & \checkmark &                           &                            & Sent                             & Many                      & WD,WP                       &                              & Date,Str,Val & Partial                           &    Wikilinks                             \\
\hline
\end{tabular}
}
\end{table}

\begin{figure}[ht!]
\caption{Encyclopedic Datasets Family Tree}
\label{fig:encyclo_fam_tree}
\centering
\includegraphics[width=0.6\linewidth]{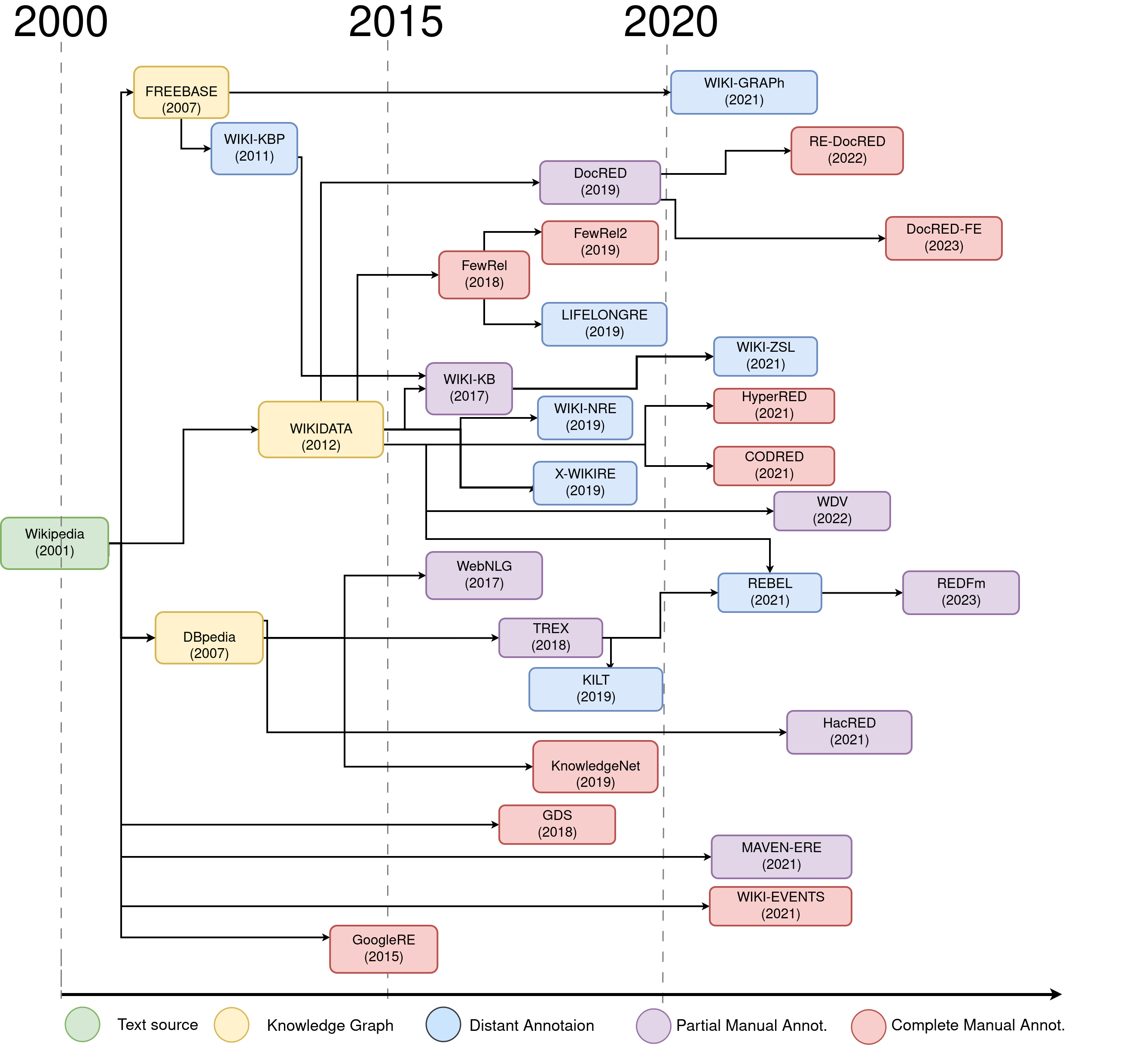}
\end{figure}
33\% of RE datasets are derived from Wikipedia, valued for its multilingual scope and rich meta-annotations such as page links. Most of these datasets are paired with Wikidata (16 of 23 cases~[\href{https://www.zotero.org/groups/6070963/scilex_re_systlitreview/collections/JDPEMF65/tags/SOURCE%3AWikidata,SOURCE%3AWikipedia/collection}{\faClipboardList}]), with additional efforts drawing on DBpedia~\cite{elsahar_t-rex_2018,bassignana_multi-crossre_2023,gardent_creating_2017} and, earlier, Freebase~\cite{noauthor_wikigraphs_2021,riedel_modeling_2010} (see Fig.~\ref{fig:encyclo_fam_tree}).
The first large-scale attempt was the \textbf{WIKI-KBP} dataset introduced by \cite{riedel_modeling_2010}, which aligned Freebase triples with Wikipedia text through distant supervision. It operated under the assumption that if a paragraph contains two entities, it potentially expresses the relation linking them in the knowledge base (KB). Entity mentions were extracted using the Stanford NER tagger, focusing on four types: \textit{Person}, \textit{Organisation}, \textit{Place} and \textit{Other}. This first proposal was followed by \textbf{GoogleRE}~\cite{noauthor_GRE_2015}, a small partially annotated collection focused on person-related relations, and later by \textbf{WIKI-KB}~\cite{sorokin_context-aware_2017}, which linked Wikipedia with Wikidata using Wikilinks for entity detection. The \textbf{WEBNLG}~\cite{gardent_creating_2017} dataset was created by verbalising (generating simple sentences expressing) triples from DBpedia using crowdsourcing. It was proposed as a benchmark for text generation from triples, but also used to evaluate text-to-triples models.

\textbf{T-REX}~\cite{elsahar_t-rex_2018} is a milestone in terms of scaling, aligning 10 million triples from DBpedia with Wikipedia content. This resource was later integrated into the KILT benchmark~\cite{noauthor_kilt_2020}, which evaluates models on multiple knowledge-intensive tasks grounded in Wikipedia content at the document level.
Subsequent efforts broadened the scope: manually annotated corpora such as \textbf{GDS}~\cite{jat_improving_2018} and \textbf{KnowledgeNet}~\cite{mesquita_knowledgenet_2019}; few-shot learning related datasets such as \textbf{FewRel}~\cite{han_fewrel_2018}, later expanded into \textbf{FewRel 2.0}~\cite{gao_fewrel_2019}, or \textbf{WIKI\-ZSL}~\cite{chen_zs-bert_2021}. Also datasets designed to solve lifelong learning as \textbf{LifelongFewRel}~\cite{wang_sentence_2019}; document-level corpora, among which \textbf{DocRED}~\cite{yao_docred_2019}, which was corrected into \textbf{RE-DocRED}~\cite{tan_revisiting_2022} and also refined into \textbf{DocRED-FE}~\cite{noauthor_docred-fe_2023}. 
More recent initiatives introduced new dimensions, including \textbf{HyperRED}~\cite{chia_dataset_2022} for \textit{n}-ary relations, textbf{CodRED}~\cite{noauthor_codred_2021} for cross-document relations, and \textbf{REBEL}~\cite{huguet_cabot_rebel_2021}, which covered over 1,000 relation types with millions of triples and led to a multilingual benchmark known as \textbf{REDFm}~\cite{cabot_redrm_2023}. 
Generative approaches also recently appeared, such as the \textbf{WDV}~\cite{amaral_wdv_2022} dataset built from Wikidata triples. 
Specialised event-extraction corpora were recently published~\cite{noauthor_maven-ere_2022,li_document-level_2021}, diversifying the landscape of available datasets.
In terms of dataset volume, resources such as \textbf{REBEL}, \textbf{T-REX}, and \textbf{CodRED} contain millions of instances, while meta-benchmarks like \textbf{KILT} or \textbf{RELD} reach $10^6$ documents. Sentence-level corpora, such as \textbf{X-WikiRE}, include up to $10^7$ short paragraphs. Encyclopedic datasets describe the largest number of relations (up to $10^4$ in RELD) and entity types (up to $10^3$ classes). By comparison, major knowledge bases such as DBpedia, Wikidata, YAGO and Freebase contained between $10^8$ and $10^9$ triples in 2016, and the English version of Wikipedia held about $10^6$ articles~\cite{färber2018knowledgegraphbestme} in 2016. \textbf{RELD} is the most recent initiative gathering previous datasets built on top of Wikipedia and covers the entire dual base of textual data from Wikipedia and triples from DBpedia.

\subsection{News Datasets}
\begin{table}[ht!]
\caption{Details on News datasets. To gain in readability, we use abbreviations to refer to tasks: NER: for Named Entity Recognition, CR: Coreference Resolution, ET: Entity Typing, EL: Entity Linking, RC: Relation Classification and GtoT: Graph-to-Text, as well as to some recurring sources: WP: Wikipedia, WD: Wikidata, DBP: DBPedia and FB: FreeBase. Also, Sent and Doc refer to sentence and of document granularity, respectively. "?" refers to cases where data is not available or clearly stated in the articles.}
\label{tab:news_ds_details}
\resizebox{\textwidth}{!}{
\begin{tabular}{|ll|lllllll|l|l|l|l|l|l|l|l|l|}
\hline
\multicolumn{1}{|l|}{\multirow{2}{*}{Year}} & \multirow{2}{*}{Short Title} & \multicolumn{7}{c|}{TASK}                                                                                                                                                               & \multicolumn{1}{c|}{\multirow{2}{*}{GRANULARITY}} & \multicolumn{1}{c|}{\multirow{2}{*}{Lang}} & \multicolumn{1}{c|}{\multirow{2}{*}{Source}} & \multicolumn{2}{c|}{Datatype}                              & \multicolumn{1}{l|}{\multirow{2}{*}{ManualAnnot}} & \multicolumn{1}{c|}{\multirow{2}{*}{NER Tools Used}} & \multicolumn{1}{l|}{\multirow{2}{*}{SynthGen}} & \multirow{2}{*}{Neg.Ex} \\ \cline{3-9} \cline{13-14}
\multicolumn{1}{|l|}{}                      &                              & \multicolumn{1}{l|}{NER} & \multicolumn{1}{l|}{CR} & \multicolumn{1}{l|}{ET} & \multicolumn{1}{l|}{EL} & \multicolumn{1}{l|}{RC} & \multicolumn{1}{l|}{RE} & \multicolumn{1}{l|}{Other} & \multicolumn{1}{c|}{}                             & \multicolumn{1}{c|}{}                      & \multicolumn{1}{c|}{}                        & \multicolumn{1}{c|}{ObjProp} & \multicolumn{1}{c|}{DTProp} & \multicolumn{1}{l|}{}                             & \multicolumn{1}{c|}{}                         & \multicolumn{1}{l|}{}                          &                         \\ \hline
2003                                        & CONLL03~\cite{tjong_kim_sang_introduction_2003}                      & \checkmark &                         &                         &                         &                         &                         &                            &                                                   & EN,DE                                      & Reuters                                      &                          & Str                         & \checkmark                                               &                                               &                                            &                     \\
2004                                        & ACE2004~\cite{doddington_automatic_2004}                      & \checkmark & \checkmark &                         &                         &                         & \checkmark & \checkmark   & Doc                                               & EN,CN,AR                                   & ?                                            & \checkmark                          & Str                         & \checkmark                                               &                                               &                                            &                     \\
2004                                        & Conll04~\cite{noauthor_cognitive_2004}                      & \checkmark &                         &                         &                         &                         & \checkmark &                            & Sent                                              & EN                                         & CognitiveCompGroup                           &                          & Str                         & \checkmark                                               &                                               &                                            &                     \\
2006                                        & ACE2005~\cite{walker_christopher_ace_2006}                      & \checkmark & \checkmark &                         &                         &                         & \checkmark & \checkmark   & Doc+Sent                                          & EN,CN,AR                                   & LDC                                          &                          & Str                         & \checkmark                                               &                                               &                                            &                     \\
2006                                        & TIMEBANK~\cite{pustejovsky_james_timebank_2006}                     &                          &                         &                         &                         &                         & \checkmark & \checkmark   &                                                   & EN                                         & LDC                                          & ?                            & Date,Str                    & \checkmark                                               &                                               &                                            &                     \\
2010                                        & NYT10~\cite{riedel_modeling_2010}                        &                          &                         &                         &                         &                         & \checkmark &                            & Sent                                              & Multi                                      & FB,NYT                                       &                          & Str                         &                                                   & StandfordCRF-NER                              &                                            &                     \\
2011                                        & WIKI-KBP~\cite{riedel_modeling_2010}                     &                          &                         &                         &                         & \checkmark & \checkmark &                            & Sent                                              & EN                                         & FB,NYT                                       & \checkmark                          & Str                         & Partial                                           & StandfordNER                                  &                                            &                     \\
2013                                        & TempEval~\cite{uzzaman_semeval-2013_2013}                     & \checkmark &                         &                         &                         &                         &                         & \checkmark   & Nsp                                               & EN,ES                                      & Aquaint,Timebank                             &                          & Date,Str                    & Partial                                           &                                               &                                            &                     \\
2014                                        & HIEVE~\cite{glavas_hieve_2014}                        &                          &                         &                         &                         & \checkmark &                         &                            &                                                   & EN                                         & GraphEVe                                     &                          & Str                         & \checkmark                                               &                                               &                                            & \checkmark                     \\
2015                                        & TAC-KBP~\cite{ellis_overview_2015}                      & \checkmark & \checkmark &                         & \checkmark &                         & \checkmark & \checkmark   & Sent                                              & EN,CN,ES                                   & Web,BaseKB,FB,NYT,Xinhua                     & \checkmark                          & Str                         & \checkmark                                               &                                               &                                            & ?                       \\
2016                                        & NYT-FB~\cite{marcheggiani_discrete-state_2016}                       &                          &                         &                         &                         &                         & \checkmark &                            & Sent                                              & EN                                    & FB,NYT                                       & \checkmark                          & Str                         &                                                   & \checkmark                                           &                                            & \checkmark                     \\
2017                                        & TACRED~\cite{stoica_re-tacred_2021}                       & \checkmark &                         & \checkmark &                         &                         & \checkmark & \checkmark   & Sent                                              & EN                                         & Tac-Kbp                                      &                          & Str                         & \checkmark                                               &                                               &                                            & \checkmark                     \\
2018                                        & NYT-MULTI~\cite{zeng_extracting_2018}                    &                          &                         &                         &                         &                         & \checkmark &                            & Sent                                              & Multi                                      & FB,NYT                                       &                          & Str                         &                                                   & StandfordCRF-NER                              &                                            &                     \\
2018                                        & MATRES~\cite{ning_multi-axis_2018}                       &                          &                         &                         &                         &                         & \checkmark & \checkmark   &                                                   & EN                                         & TBDense                                      &                          & Date,Str                    & \checkmark                                               &                                               &                                            &                     \\
2019                                        & NYT10HRL/NYT11-HRL~\cite{takanobu_hierarchical_2019}           &                          &                         &                         &                         &                         & \checkmark &                            & Sent                                              & EN                                         & FB,NYT                                       & \checkmark                          & Str                         & \checkmark                                               & StandfordNER                                  &                                            &                     \\
2020                                        & TACREV~\cite{alt_tacred_2020}                       &                          &                         &                         &                         &                         & \checkmark &                            & Sent                                              & EN                                         & Tacred                                       &                          & Str                         & Partial                                           &                                               &                                            & \checkmark                     \\
2020                                        & CASIE~\cite{satyapanich_casie_2020}                        &                          &                         &                         &                         &                         &                         & \checkmark   & Sent                                              & EN                                         & APTReppoirt,WD                               & ?                            & Str                         & \checkmark                                               & CoreNLP,DBPSpotlight                      &                                            &                     \\
2020                                        & NYT-H~\cite{noauthor_towards_2020}                        &                          &                         &                         &                         &                         & \checkmark &                            & Sent                                              & EN                                         & FB,NYT                                       & \checkmark                          & Str                         & \checkmark                                               & StandfordNER                                  &                                            & \checkmark                     \\
2021                                        & DWIE~\cite{zaporojets_dwie_2021}                         & \checkmark & \checkmark &                         & \checkmark &                         & \checkmark &                            & Doc                                               & EN                                         & Deutschewelle,WD                             & \checkmark                          & Str                         & \checkmark                                               &                                               &                                            &                     \\
2021                                        & ReTACRED~\cite{stoica_re-tacred_2021}                     &                          &                         &                         &                         &                         & \checkmark &                            & Sent                                              & EN                                         & TACRED                                       &                          & Str                         & Partial                                           &                                               &                                                & \checkmark                     \\
2021                                        & FSL TACRED~\cite{sabo_revisiting_2021}                   & \checkmark &                         & \checkmark &                         & \checkmark &                         &                            & Sent                                              & EN                                         & Tac-Kbp                                      &                          & Str                         & \checkmark                                               &                                               &                                            & \checkmark                     \\
2023                                        & RELD~\cite{pesquita_reld_2023}                         & \checkmark &                         &                         & \checkmark & \checkmark & \checkmark &                            & Doc+Sent                                          & EN                                         & Many                                         &                              & Str                         &                                                   &                                               &                                            &                         \\
2023                                        & MultiCrossRE~\cite{bassignana_multi-crossre_2023}                 & \checkmark &                         &                         &                         &                         & \checkmark &                            & Sent                                              & Multi                                      & WP,DBP                                       &                              & Str                         &                                                   &                                               & \checkmark                                            &                        \\
\hline
\end{tabular}
}
\end{table}

\begin{figure}[ht!]
\caption{News Datasets Family Tree}
\label{fig:news_ds_tree}
\centering
\includegraphics[width=.9\linewidth]{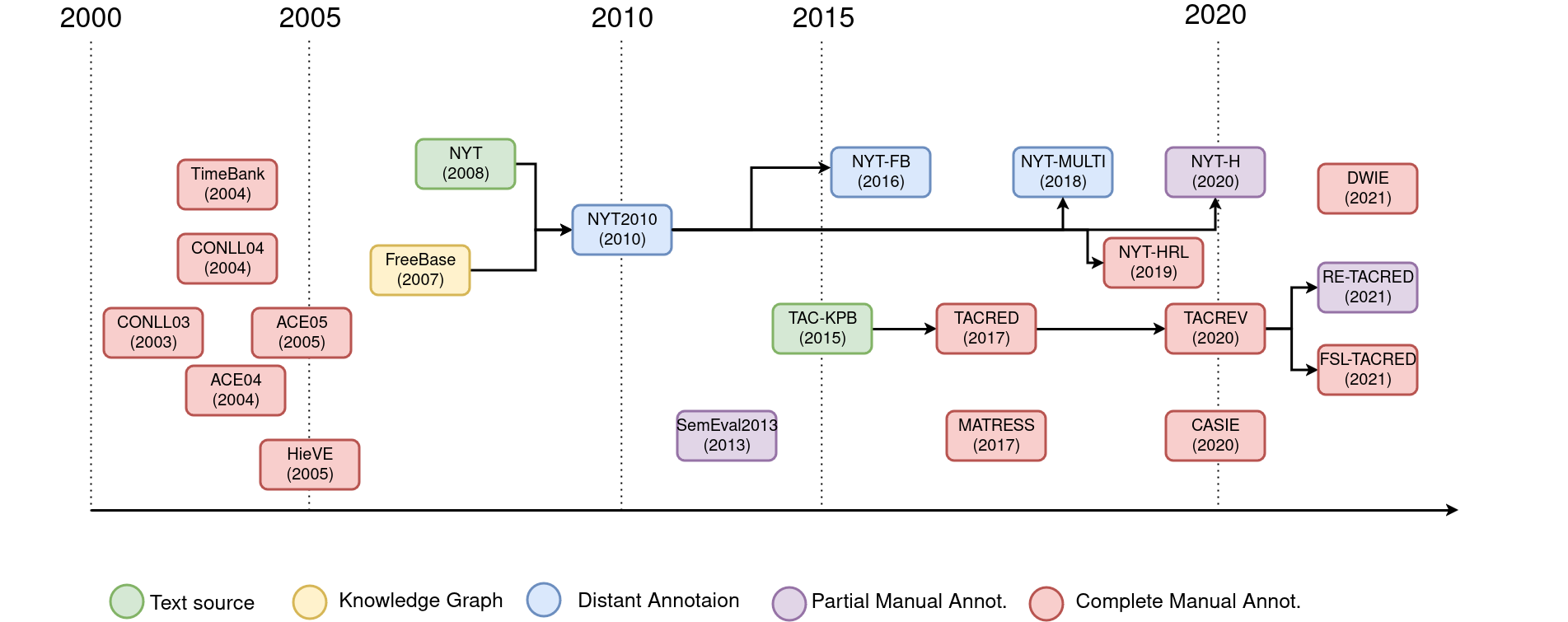}
\end{figure}

In contrast to encyclopedic datasets, news-based resources were initially developed from curated collections of news articles, fully annotated by communities of linguists. Many of these initiatives emerged from conference challenges. A prominent example is the \textit{CoNLL} shared tasks, which initially targeted the Named Entity Recognition (NER) subtask~\cite{tjong_kim_sang_introduction_2003} before expanding to include coreference resolution and relation extraction~\cite{noauthor_cognitive_2004}. Similarly, the \textbf{Automatic Content Extraction} (ACE) corpora~\cite{doddington_automatic_2004,walker_christopher_ace_2006}, produced by the \textit{Linguistic Data Consortium (LDC)}, were explicitly designed to support relation extraction research.

Due to their content, news corpora are inherently linked to events, which explains why event extraction challenges were introduced early in this domain. Notable datasets in this area include \textbf{TimeBank}~\cite{pustejovsky_james_timebank_2006} and \textbf{HiEve}~\cite{glavas_hieve_2014}, both aimed to model temporal and event-based relations. These early resources generally involved fewer than ten types of relations and entity categories. Nevertheless, the multilingual dimension was considered from the outset. This consideration was also reflected in the \textbf{ACE2004} and \textbf{ACE2005} as well in the \textbf{TAC-KBP}~\cite{ellis_overview_2015} datasets, which include English, Chinese and Arabic content.

As with encyclopedic datasets, a major leap in scalability came with the application of distant supervision. This was typically based on two canonical resources: Freebase and the \textit{New York Times (NYT)} corpus. The \textbf{NYT10}~\cite{riedel_modeling_2010} dataset was derived from these two, giving rise to a new generation of news-based relation extraction datasets. Over time, \textbf{NYT10} underwent multiple adaptations and refinements: \textbf{NYT Multi}~\cite{zeng_extracting_2018} extended the dataset to a multilingual setting; \textbf{NYT-HRL}~\cite{takanobu_hierarchical_2019} adapted it for reinforcement learning experiments; and \textbf{NYT-H}~\cite{noauthor_towards_2020} scaled and corrected the original data. One persistent challenge in this context has been the large-scale detection of entities present in Freebase, which is typically addressed using the Stanford NER toolkit.

Another landmark resource in this category was introduced by the \textit{LDC} in 2015: the \textbf{TAC-KBP} dataset. This corpus combined several sources—including the NYT, Freebase, Xinhua News, and WebBase—to meet the growing need for multilingual relation extraction. Fully annotated, TAC-KBP was subsequently extended by \textbf{TACRED} \cite{stoica_re-tacred_2021}, which in turn underwent two major corrections: \textbf{TACREV} \cite{alt_tacred_2020} and \textbf{Re-TACRED} \cite{stoica_re-tacred_2021}. These revisions highlighted the difficulty of achieving high-quality annotation and the direct impact of annotation quality on downstream usage.

Beyond relation extraction alone, news-based datasets have also supported event extraction research. Notable examples include \textbf{SemEval-2013} \cite{uzzaman_semeval-2013_2013}, \textbf{MATRES}\cite{ning_multi-axis_2018} and \textbf{CASIE} \cite{satyapanich_casie_2020}, alongside more recent contributions such as \textbf{DWIE}\cite{zaporojets_dwie_2021}, which proposes a novel set of articles finely described by the entity-centred graph from the \textit{Deutsche Welle}.

\subsection{Biology and Science-Related Datasets}

Another significant domain contributing to the production of datasets for relation extraction is the field of biology, biochemistry and biomedical research. This is a vast scientific area in which experimental work is often costly, making it highly valuable to explore and leverage existing scientific literature. An illustrative example is \textit{GenIA}~\cite{ohta_genia_2002}, the earliest dataset in our corpus, which comprises fully annotated documents designed to address multiple aspects of molecular biology relation extraction. These tasks include NER, Coreference detection and Relation Identification and Classification. 

Further initiatives, such as those reported in~\cite{li_biocreative_2016,kringelum_chemprot-30_2016}, focused on identifying the effects of chemical compounds on diseases. In these cases, relations were extracted from either medical reports or scientific articles sourced from \textit{PubMed} and enriched with the \textit{Medical Subject Headings (MeSH)} taxonomy. In five out of seven datasets, 
annotations are
provided at the document level. Regarding scale, these datasets typically cover more than $10^3$ documents and aligned triples. The largest one, \textbf{GDA}~\cite{wu_renet_2019}, contains more than $10^5$ documents with over $10^5$ triples, and was specifically designed to support deep learning models~\cite{wu_renet_2019}. 

A key characteristic of datasets in this domain is the relatively small number of relation types, generally fewer than ten per dataset. Finally, several datasets targeting more general scientific content have also been developed, such as \textit{SciREX}~\cite{jain_scirex_2020} , \textit{SemEval 2018}~\cite{gabor_semeval-2018_2018} and, more recently, \textit{SciERC}~\cite{luan_multi-task_2018}. These datasets similarly feature a limited number of predefined relations and entity types.


\section{Analysis of Papers presenting Models for the Relation Extraction Task}
Our review includes a curated collection of 104 papers presenting models for the relation extraction (RE) task and published between 2019 and 2024. This represents 51\% of our curated corpus but a very small part (3\%) of the 3,567 model papers initially collected between 2019 and 2024. As explained in Section~\ref{sec:collection}, we chose to focus on the most influential papers in order to be able to conduct an in depth analysis, rather than conducting a more superficial analysis of the entire research produced on RE models.

\textit{Granularity of Annotation:} 
Models are generally specialized for a specific granularity: 58 model papers in our corpus address RE at the sentence level only, 32 address RE at the document level only, and 13 at both levels. 
Here are the model papers dealing with document-level RE~[\href{https://www.zotero.org/groups/6070963/scilex_re_systlitreview/collections/PIBU5F3Z/tags/GRANULARITY%3ADocument/collection}{\faClipboardList}] and the those with sentence-level RE~[\href{https://www.zotero.org/groups/6070963/scilex_re_systlitreview/collections/PIBU5F3Z/tags/GRANULARITY%3ASentences/collection}{\faClipboardList}].
\textit{Language Coverage:} 
All models but one are trained on English data (103 papers~[\href{https://www.zotero.org/groups/6070963/scilex_re_systlitreview/collections/JDPEMF65/tags/LANG%3AEnglish/collection}{\faClipboardList}]). A small number extend to multilingual settings: 15 papers report training on both English and Chinese~[\href{https://www.zotero.org/groups/6070963/scilex_re_systlitreview/collections/JDPEMF65/tags/LANG%3AChinese,LANG%3AEnglish/collection}{\faClipboardList}] and, in 9 cases, additional training was conducted on Arabic datasets~[\href{https://www.zotero.org/groups/6070963/scilex_re_systlitreview/collections/PIBU5F3Z/tags/LANG%3AArabic,LANG%3AEnglish/collection}{\faClipboardList}]. The analysed model papers do not cover all the languages considered in the previously discussed multilingual data papers, like REDfm~\cite{cabot_redrm_2023} or X-WikiRE~\cite{noauthor_x-wikire_2019}. In particular, no model paper in our corpus deals with Spanish, a language that is considered by several data papers.

\begin{figure}[ht!]
\centering
\begin{minipage}{.54\textwidth}
\centering
\caption{Use of datasets by models}
\label{fig:datasets_models_use}
\includegraphics[width=\linewidth]{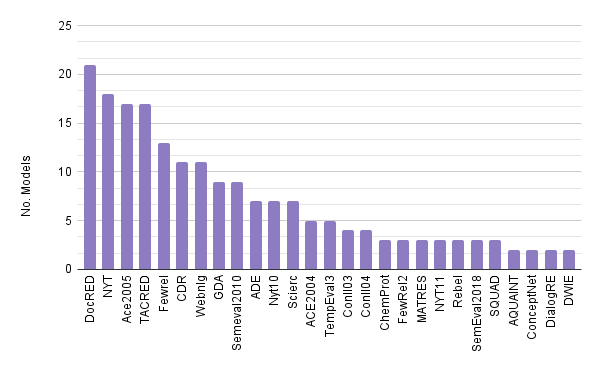}
\end{minipage}\hfill
\begin{minipage}{.44\textwidth}
\centering
\caption{Number of distinct properties managed by the proposed models }
\label{fig:property_diversity}
\includegraphics[width=\linewidth]{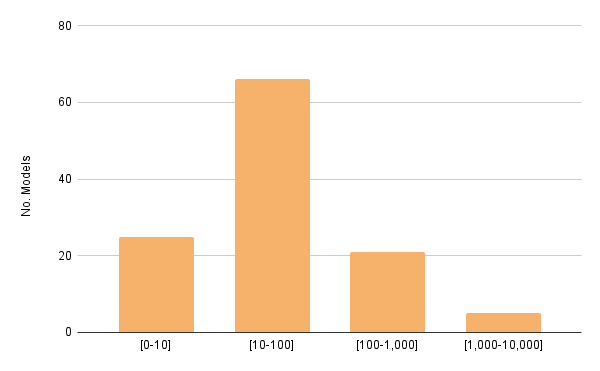}
\end{minipage}
\end{figure}

\textit{Resources Used:} 
All together, the surveyed models were trained or evaluated on approximately 100 distinct datasets. In 15 cases, models were trained on a newly created dataset developed specifically for the need of a specific proposed research question. Examples include \textbf{DocRED}, \textbf{REBEL}, \textbf{SynthIE}~\cite{josifoski_exploiting_2023} and \textbf{GenIE}~\cite{josifoski_genie_2022}. We can pinpoint from Fig.~\ref{fig:datasets_models_use} some canonical datasets that were used many times by the proposed models. As a salient example, \textbf{DocRED} was used in around 20\% of the model papers in our corpus.


\textit{Property Diversity:}
Most proposed models handle fewer than 100 relations. As shown in Figure~\ref{fig:property_diversity}, models trained and evaluated on datasets containing fewer than $10^2$ distinct relations account for more than 85\% of the studies. Models designed to manage a diversity of properties similar to those in knowledge graphs represent approximately 25\% of the proposed approaches. Only four initiatives attempt to perform on datasets with $10^3$ properties: \textbf{ERICA}~\cite{noauthor_erica_2020}, \textbf{CTN}~\cite{noauthor_constituency_2019}, \textbf{BERT-GT}~\cite{noauthor_bert-gt_2021} and \textbf{Table-Sequence}~\cite{wang_two_2020}.


\subsection{Tasks}

\begin{figure}[ht!]
\centering
\begin{minipage}{.49\textwidth}
\centering
\caption{Distribution of the tasks solved by the proposed models}
\label{fig:model_tasks}
\includegraphics[width=\linewidth]{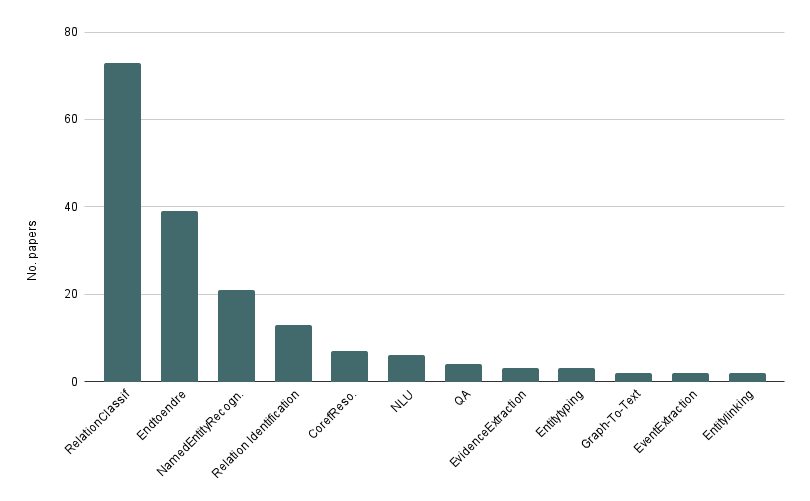}
\end{minipage}\hfill
\begin{minipage}{.49\textwidth}
\centering
\caption{Distribution of the input used by the proposed models}
\label{fig:dist_input}
\includegraphics[width=\linewidth]{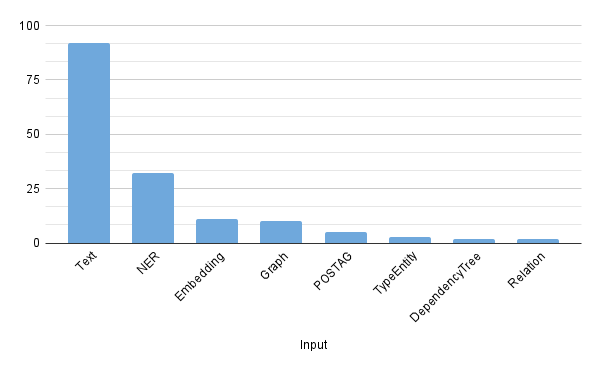}
\end{minipage}
\end{figure}

Before examining the architectural details of the proposed models, it is necessary to clarify the tasks they aim to solve. A key observation, illustrated in Figure~\ref{fig:model_tasks}, concerns the distribution of research efforts: approximately 70\% of the articles address \textit{Relation Classification}~[\href{https://www.zotero.org/groups/6070963/scilex_re_systlitreview/collections/PIBU5F3Z/tags/TASK%3ARelationClassif/collection}{\faClipboardList}], while the remaining 30\% approaches \textit{End-to-End RE models}~[\href{https://www.zotero.org/groups/6070963/scilex_re_systlitreview/collections/PIBU5F3Z/tags/TASK%3AEndToEndRE}{\faClipboardList}].  Other tasks include: \textit{Named Entity Recognition} (21 papers~[\href{https://www.zotero.org/groups/6070963/scilex_re_systlitreview/collections/PIBU5F3Z/tags/TASK%3ANER/collection}{\faClipboardList}]), \textit{Relation Identification} (13 papers~[\href{https://www.zotero.org/groups/6070963/scilex_re_systlitreview/collections/PIBU5F3Z/tags/TASK%3ARelationIdentification/collection}{\faClipboardList}]), \textit{Coreference Resolution} (7 papers~[\href{https://www.zotero.org/groups/6070963/scilex_re_systlitreview/collections/PIBU5F3Z/tags/TASK%3ACoref/collection}{\faClipboardList}]), \textit{Question Answering}~\cite{noauthor_luke_2020,noauthor_erica_2020,zhao_asking_2020,noauthor_coreferential_2020}, \textit{Entity Typing}~\cite{li_document-level_2021,noauthor_luke_2020,noauthor_erica_2020}, \textit{Graph-to-Text Generation}~\cite{dognin_regen_2021,noauthor_label_2021}, and \textit{Entity Linking}~\cite{noauthor_injecting_2021,noauthor_span-level_2019}.

We also note the emergence of tasks related to \textit{Evidence Extraction}, explored in three papers: \textbf{DREEAM}~\cite{noauthor_dreeam_2023}, \textbf{SIRE}~\cite{noauthor_sire_2021}, and \textbf{Eider}~\cite{noauthor_eider_2022}. Finally, several marginal but noteworthy tasks appear in the literature, such as \textit{Machine Reading Comprehension} (MRC)~\cite{zhao_asking_2020}, \textit{Reasoning}~\cite{zhou_claret_2022}, \textit{implicit relation detection}~\cite{zhao_infusing_2023}, and \textit{Event Extraction}~\cite{li_document-level_2021,noauthor_selecting_2022}.

\subsection{Input}

From the input perspective, the majority of RE models take raw text as input, although in some cases, embeddings of the text are used directly. Entity information is explicitly integrated into the input in 32 papers~[\href{https://www.zotero.org/groups/6070963/scilex_re_systlitreview/collections/PIBU5F3Z/tags/ARCHI%3AENtityEmbedding/collection}{\faClipboardList}], often together with \textit{Entity Type} annotations~\cite{noauthor_improving_2021,noauthor_frustratingly_2020,noauthor_label_2021}. 
In five works, \textit{Part-of-speech (POS)}–tagged text was employed as input~\cite{noauthor_recurrent_2020,noauthor_coreferential_2020,noauthor_learning_2019,noauthor_graphrel_2019,noauthor_beyond_2019}. \textit{Relation Embedding} was also incorporated as input in some approaches, particularly in graph-based architectures or knowledge-augmented strategies. Some models further extend the input space by including \textit{Relation description} directly in prompts~\cite{chen_zs-bert_2021}, or by encoding entities with additional probabilistic information, as in~\cite{noauthor_multi-view_2021}. More complex forms of auxiliary input have also been proposed, such as symbolic rules~\cite{noauthor_learning_2021}, images, or structured web content~\cite{noauthor_zeroshotceres_2020}.


\subsection{Architecture}

\begin{figure}[ht!]
\centering
\begin{minipage}{.5\textwidth}
\centering
\caption{Streamgraph of architecture families}
\label{fig:archi_streamgraph}
\includegraphics[width=\linewidth]{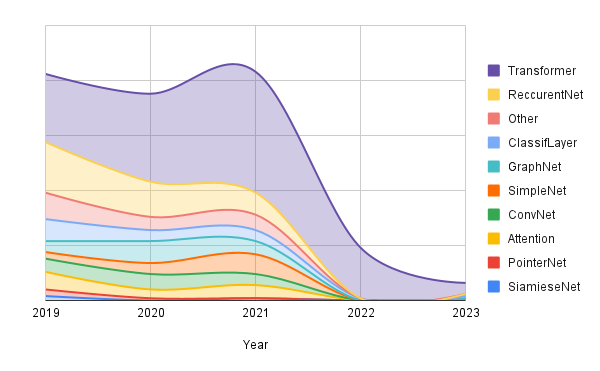}
\end{minipage}\hfill
\begin{minipage}{.5\textwidth}
\centering
\caption{Type of transformer architecture used by the models}
\label{fig:model_archi1}
\includegraphics[width=\linewidth]{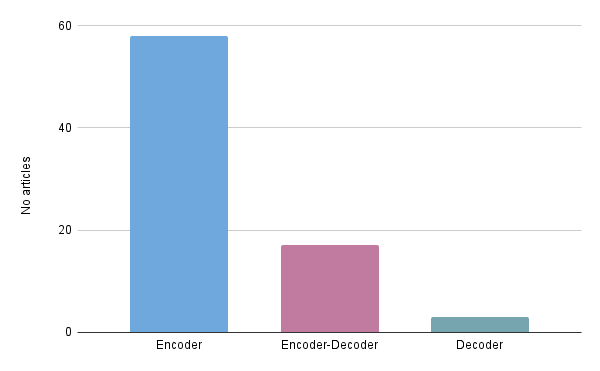}
\end{minipage}
\end{figure}

During our study, we identified 78 distinct architectural elements, reflecting the breadth of research proposals since 2019. Notably, more than 40 architectures were used only once. As illustrated in Figure~\ref{fig:model_archi1}, the introduction of Transformer-based models was followed by a dramatic surge in adoption, with 55 papers published in 2021 alone—accounting for nearly 50\% of the models in our corpus. Nevertheless, the model families highlighted in~\cite{Nayak_2021} are still represented, although their prevalence shows a progressive decline across the studied period.

Overall, these architectures can be grouped into two broad families. 
(1)~\textbf{Encoder-based models}: these approaches generally rely on a representation embedding obtained using a deep learning method, which is used as support to solve one subtask of relation extraction. In the papers studied, relation extraction with encoders relates to the relation classification subtask, where entities are already identified in a text and models have to recognise the relations expressed between them.
(2)~\textbf{SeqToSeq models}: this family includes encoder-decoder and decoder only transformer architectures, and addresses relation extraction using an \textit{End-to-End} paradigm, i.e. by directly generating a sequence of text representing the result of the extraction. 

\begin{figure}[ht!]
\centering
\begin{minipage}{.50\textwidth}
\centering
\caption{Pretrained embeddings used in models}
\label{fig:model_embed}
\includegraphics[width=\linewidth]{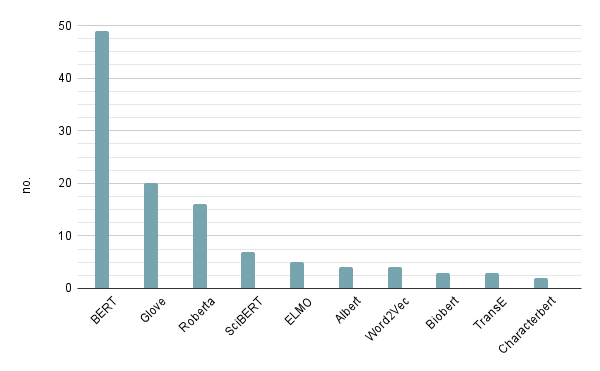}
\end{minipage}
\begin{minipage}{.45\textwidth}
\centering
\caption{Specialised embeddings created}
\label{fig:model_specembed}
\includegraphics[width=\linewidth]{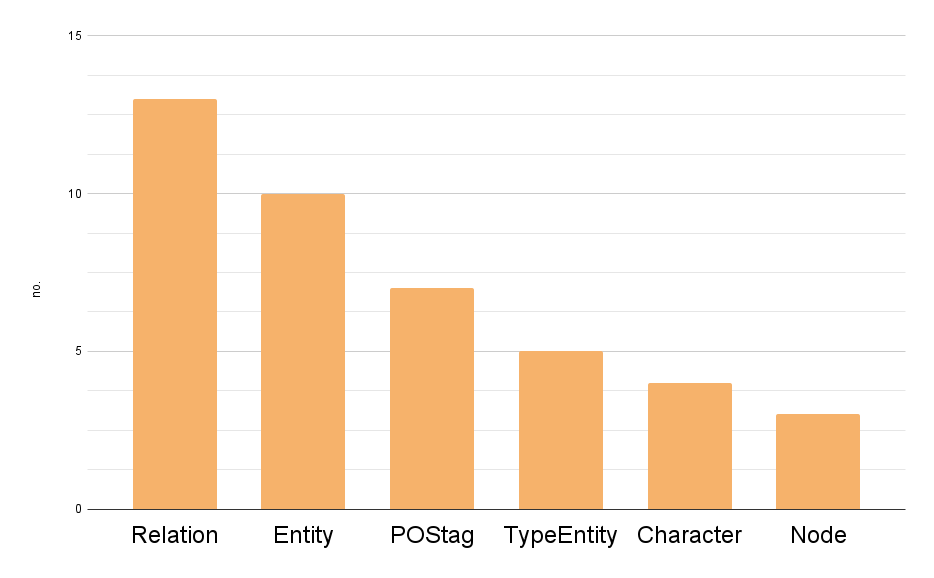}
\end{minipage}
\end{figure}


\textit{Embeddings:} 51 models in our corpus integrate or rely on embedding methods. Among the earlier approaches, we observed four papers that employed \textit{Word2vec}~\cite{noauthor_modeling_2020,nayak_effective_2020,noauthor_document-level_2019,pang_deep_2019} and approximately 20 that used \textit{GloVe}~[\href{https://www.zotero.org/groups/6070963/scilex_re_systlitreview/collections/PIBU5F3Z/tags/PTM%3AGlove/item-list}{\faClipboardList}]; however, both methods were gradually abandoned after 2020. Graph embeddings, such as \textit{TransE}, were used sparingly, appearing in only three papers~\cite{christou_improving_2021,noauthor_curriculum-meta_2021,trisedya_neural_2019}, between 2019 and 2021. Since the introduction of \textit{BERT}, most recent researches on RE have adopted BERT-related embeddings: 13 papers used the original BERT embeddings~[\href{https://www.zotero.org/groups/6070963/scilex_re_systlitreview/collections/PIBU5F3Z/tags/PTM%3ABERT/item-list}{\faClipboardList}], while others relied on BERT variations such as \textit{RoBERTa} (16 papers~[\href{ https://www.zotero.org/groups/6070963/scilex_re_systlitreview/collections/PIBU5F3Z/tags/PTM%3ARoberta/item-list}{\faClipboardList}]), \textit{ALBERT}~\cite{noauthor_label_2021,noauthor_partition_2021,wang_two_2020,noauthor_frustratingly_2020}, or more specific ones: \textit{SciBERT}~[\href{https://www.zotero.org/groups/6070963/scilex_re_systlitreview/collections/PIBU5F3Z/tags/PTM%3ASciBERT,PTM%3AScibert/item-list}{\faClipboardList}], or \textit{CharacterBERT}~\cite{noauthor_copymtl_2019,zeng_learning_2019}.

Several models also attempted to specialize embeddings to make them more effective for the RE task. For instance, three studies explored \textit{Character-level} embeddings, using as a pretrained model: \textit{ELMo}~\cite{noauthor_span-level_2019}, \textit{Word2Vec}~\cite{nayak_effective_2020}, or \textit{GloVe} ~\cite{noauthor_injecting_2021}. New representations were also investigated, generally on top of \textit{BERT} or \textit{GloVe}, including \textit{Relation Embedding} also derived from \textit{TransE}, as well as \textit{Entity-focused} embeddings, \textit{Entity Type} and \textit{Part-of-speech} (POS) features.

\textit{Recurrent Networks:}  
Recurrent networks were heavily employed before the advent of Transformer-based architectures. The simplest variant, \textit{RNN}, appeared in two papers~\cite{wang_two_2020,noauthor_learning_2019}, followed by the use of \textit{Bi-RNN} in 2019~\cite{zeng_learning_2019}. \textit{LSTM}, which is more effective at modelling long-range dependencies, was adopted in 15 papers~[\href{https://www.zotero.org/groups/6070963/scilex_re_systlitreview/collections/PIBU5F3Z/tags/ARCHI%3ALSTM/item-list}{\faClipboardList}], while \textit{GRU} was used in only three cases~\cite{noauthor_progressive_2021,noauthor_recurrent_2020,noauthor_attention_2019}. The bidirectional extension of LSTM (\textit{Bi-LSTM}) proved particularly influential, being used in 26 papers~[\href{https://www.zotero.org/groups/6070963/scilex_re_systlitreview/collections/PIBU5F3Z/tags/ARCHI%3ABILSTM/item-list}{\faClipboardList}], thus confirming the effectiveness of bidirectional designs, taking into account both side of the context during the encoding.
More advanced variants also emerged, such as hierarchical Bi-LSTMs for lifelong RE~\cite{noauthor_meta-learning_2019}. Additionally,~\cite{noauthor_extracting_2021} also proposes the use of hyperbolic layers to better capture the specific typology of relational data.

\textit{Convolutional Networks:}  
Given that graph data can be naturally represented as matrices, convolutional approaches were also widely adopted, with 12 papers designing the RE task using \textit{CNNs}~[\href{https://www.zotero.org/groups/6070963/scilex_re_systlitreview/collections/PIBU5F3Z/tags/ARCHI%3ACNN/item-list}{\faClipboardList}]. Variants such as \textit{Piecewise Convolutional Neural Networks} (PCNNs) were also investigated to better focus on important input content~\cite{noauthor_revisiting_2020,noauthor_distant_2019,noauthor_are_2019}, alongside the use of specialized architectures such as \textit{U-Net}, as reported in one ppaper~\cite{noauthor_document-level_2021}. Logical reasoning modules also emerged, such as the use of \textit{R-GCN} in the \textit{SIRE} model~\cite{noauthor_sire_2021}. 

\textit{Attention Layers:}  
Attention mechanisms were explicitly incorporated in 10 papers~[\href{https://www.zotero.org/groups/6070963/scilex_re_systlitreview/collections/PIBU5F3Z/tags/ARCHI%3AAttention/item-list}{\faClipboardList}], highlighting their growing role as standalone architectural components, even before they became dominant in transformer-based models. In the details, some proposals specialised the mechanism to better fit the input/output specificity, using ~\textit{Graph Attention}~\cite{noauthor_zeroshotceres_2020},~\textit{Hierarchical Attention}~\cite{noauthor_distant_2019}, \textit{N-gram Attention}~\cite{trisedya_neural_2019} or ~\textit{Selective Attention} strategy~\cite{noauthor_fine-tuning_2019}.

\textit{Structured Networks:}  
Several models proposed specific structured architectures, including \textit{Span Graph Tokens}~\cite{noauthor_entity_2019}, \textit{Gaussian Graph Generators} (GDPNet) integrated within \textbf{BERT-GT}~\cite{noauthor_bert-gt_2021}, and a \textit{Structured Perceptron} for event extraction~\cite{noauthor_structured_2019}. Other examples include prototype-based learning such as \textit{ProtoNet}~\cite{noauthor_learning_2020}.  

\textit{Contrastive Networks:}  
Also discussed latter, contrastive approaches were explored to better handle negative examples, and generally imply specific training processes. For instance, \textit{Siamese Networks}~\cite{noauthor_extracting_2021} and \textit{GAN-based models} (e.g., \textit{CopyMTL}~\cite{noauthor_copymtl_2019}) were proposed to enhance robustness and discriminative power.  

\textit{Copy Mechanism:}  
Finally, several specialized network designs were introduced. For example, \textit{Pointer Networks} were adopted several times~\cite{nayak_effective_2020,zeng_learning_2019,noauthor_copymtl_2019,pang_deep_2019}  to implement copy-paste mechanisms from input to output.


\textit{End-To-End models:}  Next to the embeddings appraoches, End-to-End models constitute 20\% of the models in our corpus, all of which relying on pretrained transformers. Among them, a significant number are Encoder-Decoder models, notably based on \textit{BART} (11 papers~[\href{https://www.zotero.org/groups/6070963/scilex_re_systlitreview/collections/PIBU5F3Z/tags/PTM%3ABART/item-list}{\faClipboardList}]) and ~\textit{T5}~\cite{lu_unified_2022,noauthor_exploring_2022,noauthor_dore_2022,dognin_regen_2021}, as well as more specialized versions such as \textit{BioBART}~\cite{noauthor_dore_2022} and \textit{SciFive} ~\cite{noauthor_exploring_2022}. \textit{FLAN-T5} was also used in~\cite{josifoski_exploiting_2023}. On the decoder side, there are initiatives using decoder-only pretrained models such as \textit{GPT-2}~\cite{noauthor_fine-tuning_2019} and \textit{GLM} ~\cite{noauthor_deepstruct_2022}, alongside emerging research using \textit{GPT-3.5}. 
Although decoder-only models are today the most widely adopted design in large language models (LLMs), the collected papers make little use of them. The models covered by our study and dealing with LLMs use them as synthetic data generators rather than as relation extractors.


\subsection{Adaptation to the RE Task}

\textit{Usage of Negative Examples:} Half of the analysed models~[\href{https://www.zotero.org/groups/6070963/scilex_re_systlitreview/collections/PIBU5F3Z/tags/USENEGATIVEEXAMPLE_BIN%3A1/item-list}{\faClipboardList}] use negative examples. They are either encoder-based models addressing RE as a classification task or seq2seq models updating the Loss function by taking into account these negative examples. 

\textit{Usage of synthetic data:} Four proposals integrate synthetic data into their framework. For example, \cite{li_document-level_2021} generate in a first step a summary containing the most important information of the input in order to extract, in a second step, the relation from this summary.
\cite{noauthor_label_2021} and \textbf{RelationPrompt}~\cite{noauthor_relationprompt_2022} use templates crafted for several relations to compose synthetic sentences used as support of the method proposed. Finally, \textbf{SynthIE} was trained thoroughly on synthetic text generated using triples coming from \textit{Wikidata}. 

\begin{figure}[ht!]
\centering
\begin{minipage}{.44\textwidth}
\centering
\caption{SeqToSeq Pretrained models used in the papers}
\label{fig:seqtoseq_prenained}
\includegraphics[width=\linewidth]{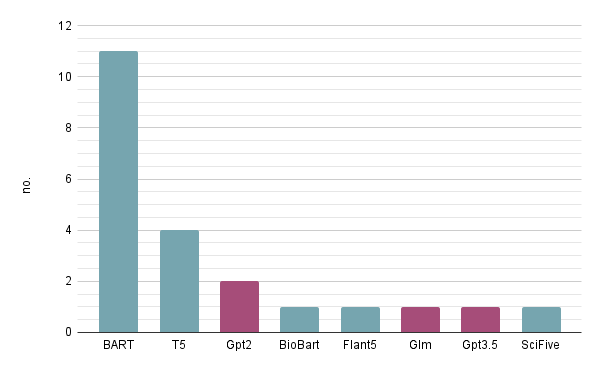}
\end{minipage}\hfill
\begin{minipage}{.56\textwidth}
\centering
\caption{Learning strategies distribution}
\label{fig:strat_distrib}
\includegraphics[width=\linewidth]{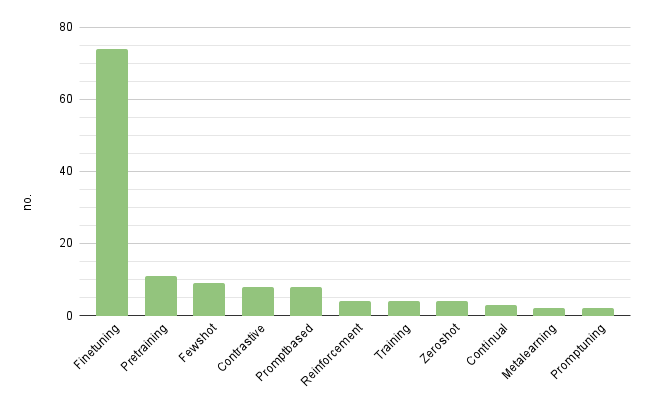}
\end{minipage}
\end{figure}

\textit{Training:} Figure~\ref{fig:strat_distrib} shows that \textit{Fine-tuning} is employed in 70\% of the model papers to address the RE task using the previously described architectures, allowing pretrained models or embeddings to be specialized. Alternatively, in 10\% of the cases, researchers continue the \textit{pretraining}, i.e. extend the training of a pretrained transformer model (PTM) using new types of textual content and, optionally, new pretraining objectives. 
Less common are initiatives training models \textit{from scratch}, with only examples re-training from scratch the embeddings for encoder-based models. As noted in previous sections, models can be specifically trained to recognize negative examples using \textit{Contrastive architectures}~[\href{https://www.zotero.org/groups/6070963/scilex_re_systlitreview/collections/PIBU5F3Z/tags/LEARNINGMETHOD%3AContrastive/item-list}{\faClipboardList}], which require dedicated training techniques. Additionally, \textit{Reinforcement learning} is a valuable tool to better align the model's behaviour with user expectations. It is used in three papers of our corpus~\cite{noauthor_selecting_2022,dognin_regen_2021,zeng_learning_2019}. 

To assess the generalizability of models to unseen knowledge, researchers increasingly explore \textit{Prompt-based}~[\href{https://www.zotero.org/groups/6070963/scilex_re_systlitreview/collections/PIBU5F3Z/tags/LEARNINGMETHOD%3APromptbased/item-list}{\faClipboardList}] and \textit{Instruction-based Learning}~\cite{lu_unified_2022}, where input text is structured to define the task explicitly or to integrate more context. Another option to reduce the computational cost is to fine-tune the models using a reduced number of additional layers related to the prompt input. This strategy is also called \textit{Prompt Tuning}~\cite{ma_prompt_2022,chen_knowprompt_2022} and can be considered as a specific kind of \textit{Parameter-Efficient Fine-Tuning (PEFT)}~\cite{zhao_infusing_2023}.

In this line, \textit{Few-shot}~[\href{https://www.zotero.org/groups/6070963/scilex_re_systlitreview/collections/PIBU5F3Z/tags/LEARNINGMETHOD%3AFewshot/item-list}{\faClipboardList}] and \textit{Zero-shot}~[\href{https://www.zotero.org/groups/6070963/scilex_re_systlitreview/collections/PIBU5F3Z/tags/LEARNINGMETHOD%3AZeroshot/item-list}{\faClipboardList}] evaluation and training have emerged as important approaches. To accommodate model updates, \textit{Continual learning} (also referred to as Lifelong learning or meta-learning) is applied in five papers~[\href{https://www.zotero.org/groups/6070963/scilex_re_systlitreview/collections/PIBU5F3Z/tags/LEARNINGMETHOD%3AContinual/item-list}{\faClipboardList}], enabling models to adapt over time. More marginally, other techniques have been employed, including expectation-maximization optimization~\cite{noauthor_learning_2021}, self-training~\cite{hu_semi-supervised_2021} or snowballing~\cite{noauthor_neural_2019}, and teacher-forcing in \textbf{DREEAM} approaches were conducted.

\textit{Loss Update:} The choice of the loss function is a key element of the configuration used to build RE models. 
70\% of the models define a specific loss function to optimize the proposed architectures. This is particularly common for encoder-based models and occurs less frequently with end-to-end models. 
For encoder-only models, an example is provided by~\cite{noauthor_fine-tuning_2019}, while for encoder-decoder and decoder architectures, ClarET~\cite{zhou_claret_2022} demonstrates tailored loss function design to improve performance.

\textit{Decoding Method:} End-To-End models generally allow the definition of a decoding strategy; many of them use beam search as the primary generation strategy, which is usually the default configuration of a PTM, as done in \textbf{REBEL}~\cite{huguet_cabot_rebel_2021}. \textbf{GenIE}~\cite{josifoski_genie_2022} introduces a constrained beam search combined with prefix trees, allowing constraints generation over the \textit{Wikidata} schema. Additionally, \textbf{DORE}~\cite{noauthor_dore_2022} uses token-ID vocabulary constraints during decoding to guide generation.

\subsection{Output}

\textit{Linearization:} 
Seq2seq models are capable of directly generating a sequence of tokens, the strategy chosen to represent a graph is generally referred to as \emph{Linearization}. Several approaches have been proposed in the literature. The most frequent approaches are to represent a triple as a \textit{list} of its subject, predicate and object, enclosed in parentheses,  or to use tags in square brackets to delimit the subject, property and object of the triple. The definition of the delimiter tokens is done before the fine-tuning, and their representation are learnt during it. 
There are also approaches representing triples in a syntax close to that of first-order logic. Other approaches factorize the triples representation to avoid repetition of subjects and, in some cases, predicates. Additionally, a few approaches use more complex syntaxes like JSON.

\begin{table}[ht!]
\caption{Linearisation details}
\resizebox{0.5\linewidth}{!}{
\begin{tabular}{|l|l|l|}
\hline
                                       Linearisation  & Factorised & Models \\ \hline
(s,p,o)                                & 0          & \cite{noauthor_dore_2022,noauthor_contrastive_2020,yuan_relation-specific_2020}         \\ \hline
<s,p,o>                                & 0          & \cite{josifoski_genie_2022,trisedya_neural_2019}          \\ \hline
<triplet>s<subj>p<obj>o<sub>...        & \checkmark & \cite{huguet_cabot_rebel_2021}          \\ \hline
[s]s[r]p[o]o[p]...                     & \checkmark & \cite{josifoski_exploiting_2023}         \\ \hline
p(s,o) p(s,o)                          & 0          & \cite{noauthor_exploring_2022}          \\ \hline
JsonData                               & \checkmark & \cite{rossiello_knowgl_2022}          \\ \hline
HeadEntity: s TailEntity: o Relation: p & 0          & \cite{noauthor_relationprompt_2022}          \\ \hline
(entytyType: s, (r:o),...)             & \checkmark & \cite{lu_unified_2022}          \\ \hline
\end{tabular}
}
\end{table}

\textit{Datatypes of Output Values:} In 39\% of the reviewed model papers (41)~[\href{https://www.zotero.org/groups/6070963/scilex_re_systlitreview/collections/PIBU5F3Z/tags/DATATYPEPROP%3AString/item-list}{\faClipboardList}]) the data is extracted only as string values. In 37\% of the reviewed model papers (39)~[\href{https://www.zotero.org/groups/6070963/scilex_re_systlitreview/collections/PIBU5F3Z/tags/OBJECTPROPERTIES_BIN%3A1/item-list}{\faClipboardList}]), the data is extracted only as URIs. The extraction is conducted using both literal values and URIs in only 20 models~[\href{https://www.zotero.org/groups/6070963/scilex_re_systlitreview/collections/PIBU5F3Z/tags/DATATYPEPROP%3AString,OBJECTPROPERTIES_BIN%3A1/item-list}{\faClipboardList}]. A few model papers explicitly represent dates (7) or typed numerical values (2).

\subsection{Computational Cost}

\begin{table}[ht!]
\resizebox{0.5\linewidth}{!}{
\begin{tabular}{|l|l|l|ll|ll|ll|}
\hline
\multirow{2}{*}{Year} & \multirow{2}{*}{Model} & \multirow{2}{*}{Type} & \multicolumn{2}{c|}{Finetuning} & \multicolumn{2}{c|}{Pretraining} & \multicolumn{2}{c|}{Total} \\
                      &                        &                       & min            & max           & min            & max            & min              & max              \\
                      \hline
2020                  & LUKE                   & encoder-based         & 0,16           & 3,3           & 720            & ?              & 720              & 725              \\
2020                  & BERT+MTB               & encoder-based         & 48             & ?             & 96             & ?              & 50               & 150              \\
2020                  & CorefBERT              & encoder-based         & ?              & ?             & 288            & 2112           & 300              & 2120             \\
2021                  & SIRE                   & encoder-based         & ?              &               & 18             & 24             & 20               & 30               \\
2021                  & REBEL                  & encoder-decoder       & 0,27           & 5,6           & 27             & 200            & 0,27             & 210              \\
2021                  & REDSandT               & encoder-based         & 0,15           & 5             & ?              & ?              & 0.15             & 5                \\
2021                  & BERT-GT                & encoder-based         & ?              & ?             & 5              & ?              & 5                & 10               \\
2022                  & GenIE                  & encoder-decoder       & 2,5            & 60            & 12             & 492            & 14,4             & 492               \\
2022                  & ClarET                 & encoder-decoder       & ?              &               & 90             & ?              & 92               & 100              \\
2023                  & SynthIE                & encoder-decoder       & ?              & ?             & 160            & 240            & 162              & 300              \\
2023                  & DREEAM                 & encoder-based         & 0,5            & ?             & 4              & 7              & 4,5              & 7,5          \\
\hline
\end{tabular}
}
\caption{Estimated computational cost in GPU hours, based on the information available and detailed in the model papers. "?" refers to cases where data is not available or clearly stated in the papers.}
\label{fig:GPU_cost_comparison}
\end{table}

In the entire analysed corpus only 11 model papers clearly detail and discuss the computational cost of their experiments. As noted in the previous section, models may involve both pretraining and fine-tuning steps. The cost of these processes is reported in various units across the literature, including GPU days, GPU hours or GPU minutes, and must also account for the number of GPUs used. Comparing these results is challenging because recorded times depend not only on the dataset size but also on the GPU hardware employed.  

We collected this information and estimated the minimum and maximum costs of each model, as summarised in Table~\ref{fig:GPU_cost_comparison}. Pretraining is generally the most costly strategy compared to fine-tuning. Encoder-based architectures typically require smaller budgets, essentially because encoder-decoder models often demand more resources for pretraining, which is usually performed before fine-tuning on specific evaluation datasets. From the encoder perspective, \textbf{CorefBERT} provides an example of a model pretrained from scratch, which requires even more computational resources.


\section{Citation network}
\begin{figure}[ht!]
\centering
\includegraphics[width=0.8\textwidth]{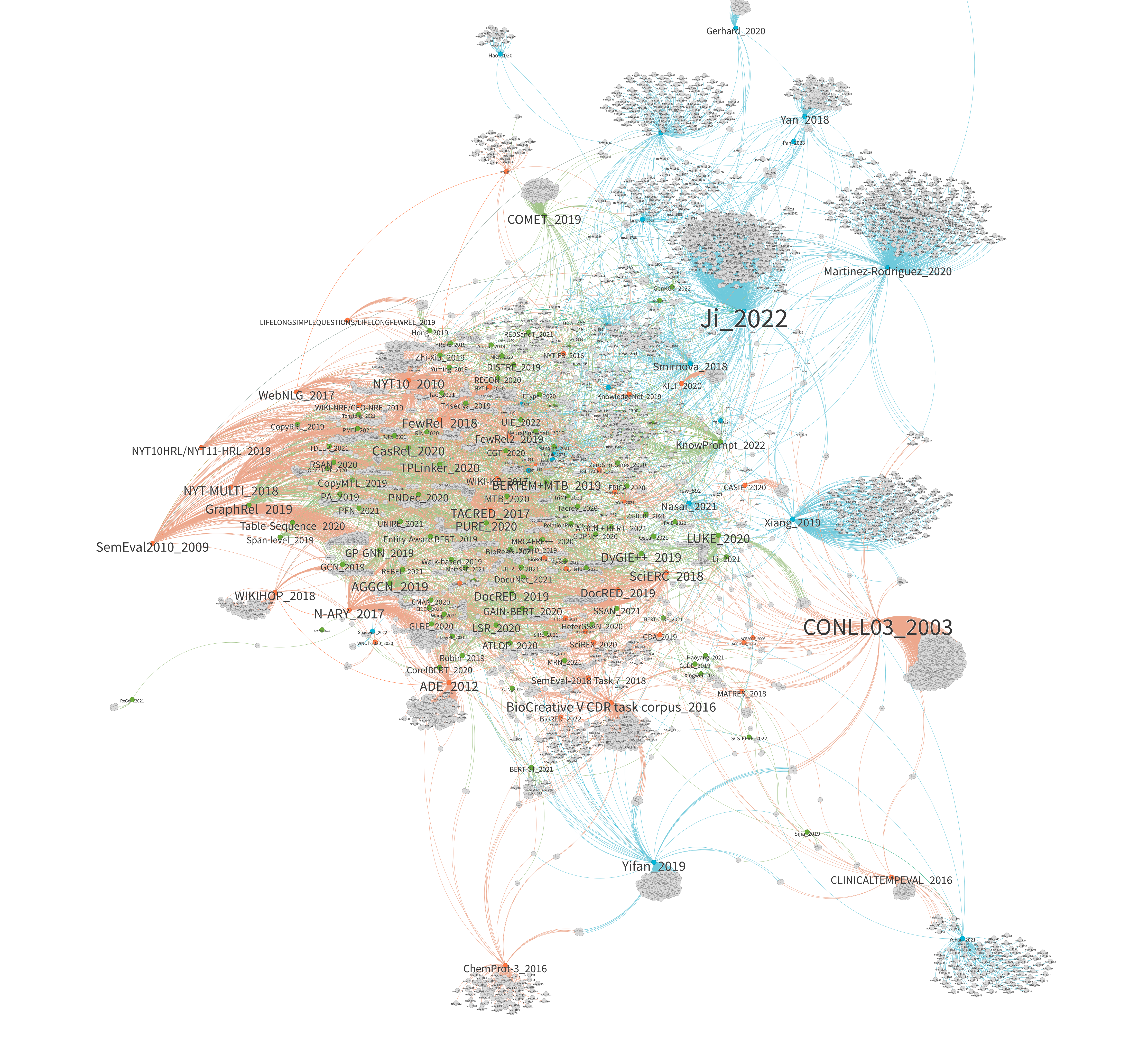}
\caption{Citation network, with model papers in green, dataset papers in blue, surveys in orange and papers out of the scope in grey. The color of an edge is the combination of the colors of the nodes it links.}
\label{fig:scilexnetwork}
\end{figure}

\textit{Global Overview:} In our collection of 202 papers, 190 papers are identified by a DOI enabling to retrieve their citation network. The complete network was expanded to 6,253 papers by incorporating all the papers identified in the citation data of our corpus. The resulting graph has an average degree of 1.63 and a density close to zero, indicating that it is extremely sparse. This sparsity reflects a common pattern in citation networks, where most papers cite only a small subset of all the available works. A visualisation of the network is presented in Fig.~\ref{fig:scilexnetwork}\footnote{\href{https://datalogism.github.io/SciLexRE/index.html}{Interactive version available online}}. Despite its sparsity, the graph has a diameter of 7, meaning that the longest shortest path between any two papers spans seven citation steps. This observation is consistent with the “small-world” property typically associated with citation networks, where citation chains connect otherwise distant works. The analysis of the network’s community structure yields a modularity score of 0.7, a relatively high value that indicates well-defined clusters of papers. Conversely, the clustering coefficient is low and close to zero, which can be explained by the fact that papers tend to cite prior work, yet two papers citing the same work have a relatively low probability of citing each other.

\textit{Analysis by paper type:} A closer inspection highlights the central role and strong impact of survey papers in the network. Surveys summarize the state of the art and frequently serve as reference points for subsequent research. Notable examples include the comprehensive survey by~\cite{9416312}, which provides a mathematically grounded overview of embedding methods and models for knowledge representation; the benchmarking initiative in the biomedical domain by~\cite{peng-etal-2019-transfer}; and the influential work of~\cite{8918013} addressing event extraction. Other key contributions include~\cite{martinez-rodriguez_information_2020} and~\cite{Yan2018ARO}, both of which focus on knowledge graph representation. Dataset papers also occupy central positions in the network, often anchoring specific communities of practice. Examples include the foundational CoNLL03 dataset~\cite{tjong_kim_sang_introduction_2003}, as well as the hubs represented by CDR~\cite{li_biocreative_2016} and ADE~\cite{gurulingappa_development_2012}. Models proposed by the community are generally concentrated near the network’s core, while more specialized or innovative approaches tend to be positioned towards the periphery.

\textit{Communities analysis:} With the Louvain community detection algorithm, we identified over 40 clusters, highlighting the diverse research communities involved in the RE task, typically centred on specific datasets. The largest cluster is drawn by~\cite{9416312}, which serves as a key foundation for encoder-based research. Another significant cluster encompasses news and event-related models, including FewRel, which has later been used to benchmark models trained on news datasets. We observed a notable cluster around the CoNLL03 dataset, frequently cited by many papers as one of the first datasets for Named Entity Recognition (NER) tasks. Additionally, there is substantial research activity surrounding the NYT-10 and NYT-H datasets, which paved the way for distant annotation techniques, along with updated versions like NYTHT10 and NYT11, exemplified by works like REBEL, TPLinker and RSAN. Fianlly, there is a cluster related to the N-Ary and WikiHop datasets, which tackle complex relations and have inspired various convolutional neural network approaches. 

\section{Benchmark}

\textit{Metrics:} In the literature, a variety of metrics were used to evaluate relation extraction models, depending on the task. \textbf{Ranking metrics} such as Hits@k, Mean Reciprocal Rank (MRR) and Mean Rank (MR) are commonly applied in knowledge graph completion tasks to assess the rank of correct predictions. \textbf{Classification metrics}, including classification Accuracy, Micro/Macro F1 and Ignored F1 (Ign.F1) measure the correctness of predicted relations, with some metrics ignoring irrelevant labels. \textbf{Curve-based metrics} like the Area Under the Curve (AUC) evaluate model performance across different decision thresholds. For sequence generation approaches, \textbf{text generation metrics} such as BLEU and ROUGE assess the quality of generated sequences by comparing them to reference triples or linearized outputs. But the most commonly used metric remains the micro F1, which is used in the following to compare and evaluate the performances of the analysed models of the corpus. 

\begin{figure}[ht!]
\begin{minipage}[c]{\textwidth}
\centering
    \includegraphics[width=.6\linewidth]{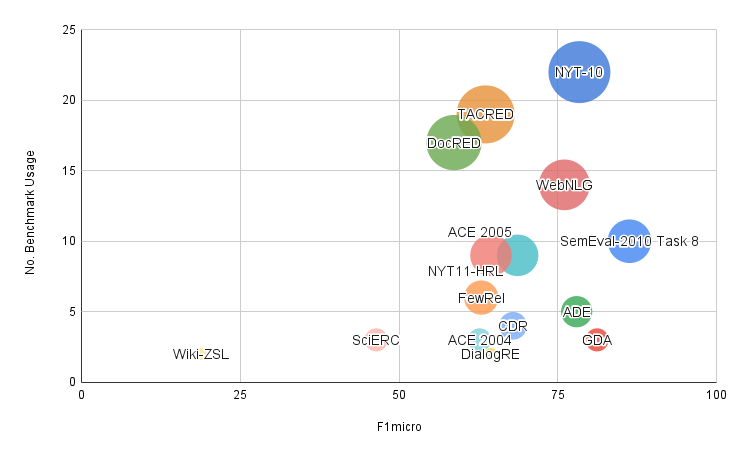}
    \caption{Datasets usage and averaged micro F1 results}
    \label{fig:dataset_use_F1}
\end{minipage}
\end{figure}

\textit{Dataset benchmarked:} The statistics gathered from PapersWithCode and previous surveys allow us to consolidate and analyze the results of 25 models evaluated on 29 datasets. Despite the importance of the \textit{macro F1} metric for assessing a model's ability to handle underrepresented relations, many studies report only micro-level results. Therefore we use the micro-F1 scores if available to compare all our results. Among the datasets, only 14 are used in more than two model papers, they are illustrated in Fig.~\ref{fig:dataset_use_F1}. Considering a benchmark as \textit{saturated} when at least one model achieves an F1 score higher than 0.9, we observe that \textbf{NYT-10} and \textbf{WebNLG} are saturated by six models, while \textbf{SemEval-2010 Task 8} and \textbf{Re-TACRED} are saturated by one model each. Beyond these, datasets such as \textbf{TACRED}, \textbf{DocRED}, \textbf{ACE2005}, \textbf{NYT11-HRL}, and \textbf{FewRel} remain challenging for current models.

\begin{figure}[ht!]
\centering
\begin{minipage}{.45\textwidth}
\centering
\includegraphics[width=\linewidth]{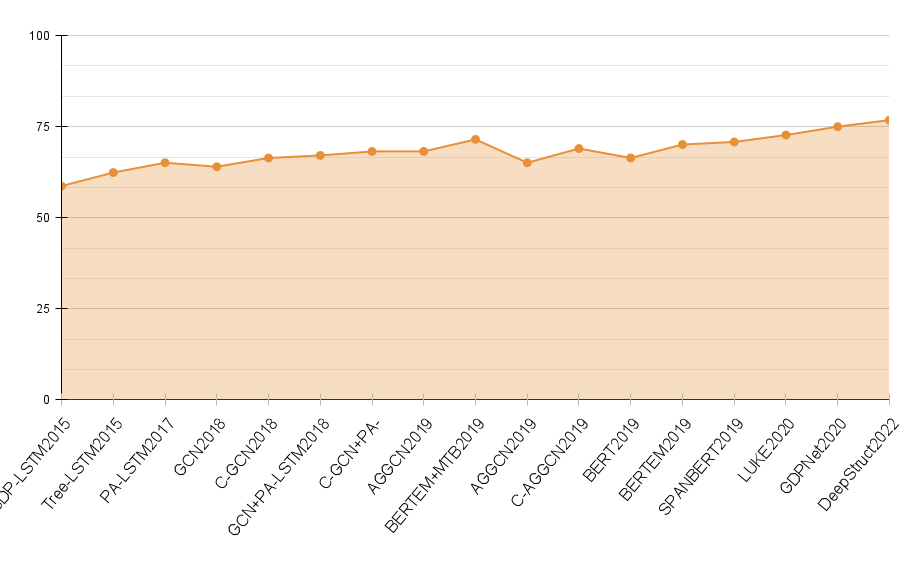}
\caption{MicroF1 on TACRED}
\label{fig:tacred_year}
\end{minipage}\hfill
\begin{minipage}{.45\textwidth}
\centering
\includegraphics[width=\linewidth]{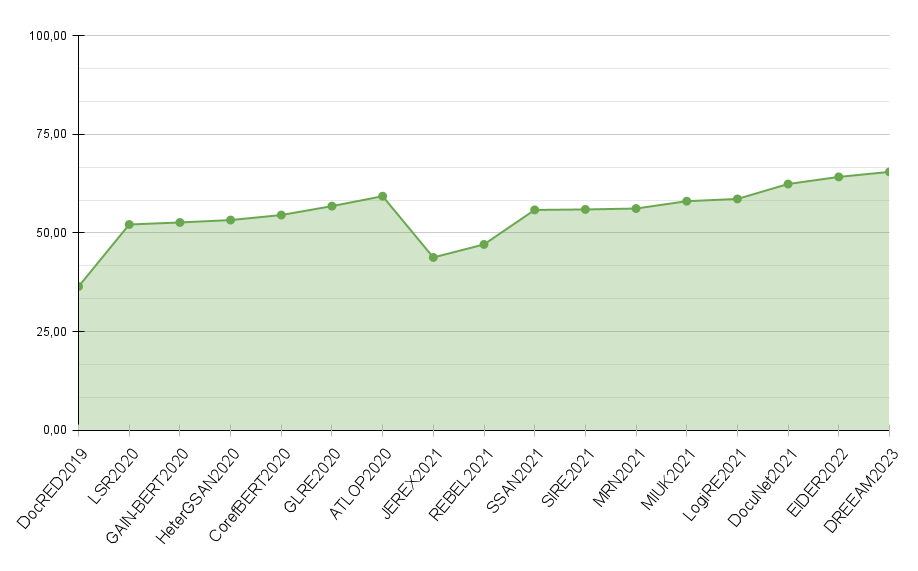}
\caption{MicroF1 on DOCRED}
\label{fig:docred_year}
\end{minipage}
\end{figure}
\begin{figure}[ht!]
\centering
\begin{minipage}{.45\textwidth}
\centering
\includegraphics[width=\linewidth]{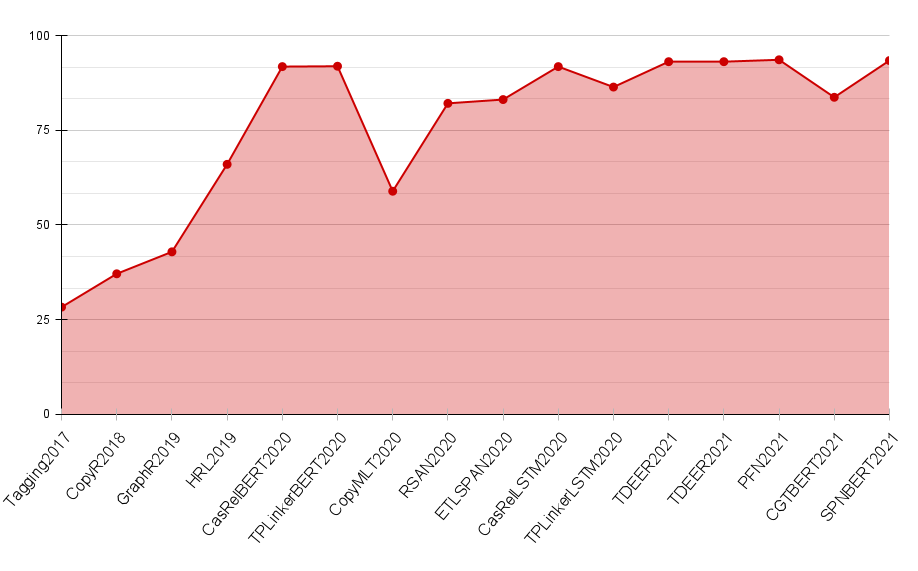}
\caption{MicroF1 on WEBNLG}
\label{fig:webnlg_year}
\end{minipage}\hfill
\begin{minipage}{.45\textwidth}
\centering
\includegraphics[width=\linewidth]{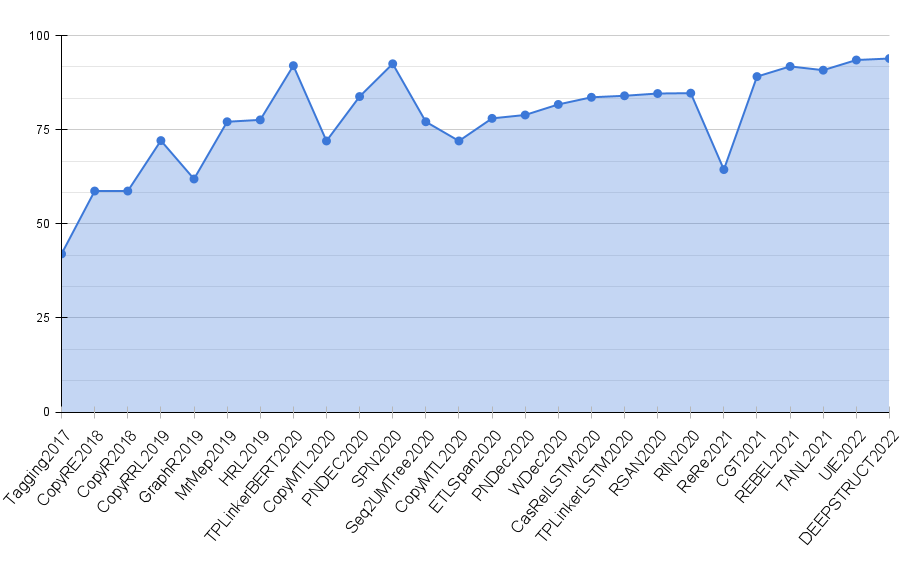}
\caption{MicroF1 on NYT}
\label{fig:nyt_year}
\end{minipage}
\end{figure}

\textit{Performance Trends:} Our analysis highlights clear performance improvements on four widely studied datasets: TACRED, DocRED, WebNLG and NYT-10 (see Fig.~\ref{fig:tacred_year},~\ref{fig:docred_year},~\ref{fig:webnlg_year},~\ref{fig:nyt_year}). Among them, the most striking progress was observed on \textbf{WebNLG}, where micro-F1 scores increased by 222\% between 2017 and 2021. This leap can be attributed to the dataset’s design: short texts containing few facts, which favour effective modelling. Comparable gains were recorded on \textbf{NYT-10}, with an improvement of 121\% between 2017 and 2022. Early  encoder-decoder systems, such as \textbf{PNDEC}~\cite{nayak_effective_2020}, achieved F1 scores around 0.8, while later seq2seq architectures (e.g., \textbf{DeepStruct}~\cite{noauthor_deepstruct_2022}, \textbf{REBEL}~\cite{huguet_cabot_rebel_2021},  \textbf{UIE}~\cite{lu_unified_2022}) consistently reached or exceeded these levels. Notably, \textbf{NYT-10} exhibited earlier improvements than other benchmarks, with a 39\% relative gain as early as 2018, whereas most datasets only saw significant advances after 2020. Its other version, \textbf{NYT11}, has been less frequently benchmarked; however, BERT-based approaches still push F1 scores close to 0.9, compared to pre-2019 baselines of around 0.6.

In contrast, \textbf{TACRED} and \textbf{DocRED} showed only modest improvements. On \textbf{TACRED}, the best models achieved micro-F1 scores near 0.75, compared  to pre-2019 baselines of about 0.6. \textbf{DeepStruct} achieved 0.76, while \textbf{REBEL} reported a notable 0.94 on the more robust Re-TACRED variant. DocRED remains particularly challenging: while DREEAM achieved the best F1 (0.67), most earlier systems hovered around 0.6. \textbf{REBEL} underperformed here, also around 0.6, underscoring the challenge of adapting models to long-document contexts.

Other benchmarks show similar trends. On \textbf{ACE2005}, F1 scores improved modestly from 0.60 to 0.65, while \textbf{CDR} rose from 0.60 to 0.75. \textbf{SemEval2010} tasks saw a sharper jump, from early CNN/SVM systems (\~0.80) to \textbf{KnowPrompt}~\cite{chen_knowprompt_2022} (\~0.90). In contrast, \textbf{SciERC} illustrates the persistent difficulty related to domain-specific RE, with performance plateauing below 0.60.

Few-shot datasets further highlight recent progress. Early prompt-based methods, such as \textbf{RelationPrompt}~\cite{noauthor_relationprompt_2022}, achieved an F1 score of only 0.20 in 5–10 shot scenarios. In contrast, recent models, such as \textbf{DeepStruct}, nearly saturated the benchmark, achieving near-perfect extraction in both 5- and 10-shot settings.

\section{Trends and the Story so Far}
Since its formalization in the early \textbf{1990s}, research on relation extraction (RE) has evolved through several major shifts. The first decade relied on manually crafted patterns and statistical methods. With the advent of Wikipedia in \textbf{2001}, manually annotated corpora became available, establishing the first robust baselines, often based on newswire data. The emergence of Linked Open Data (\textbf{2007–2014}) further accelerated progress by enabling structured knowledge graphs at scale, which in turn supported the creation of large, distantly supervised datasets for training deep learning models. For much of this period, RE was framed as a pipeline task: entities were identified using NER systems and subsequently linked through classification models, increasingly supported by deep learning architectures such as LSTMs, recurrent networks, and GANs. A decisive boost came with the introduction of distributed word representations, starting with \textit{word2vec} in \textbf{2013}, which established embeddings as a foundation for modern NLP.

Our survey focuses on the more recent \textbf{Transformer era}, which has defined the last five years of RE research. This phase began with the “attention” mechanism and was consolidated in \textbf{2019} with BERT fine-tuning, applied in numerous configurations through task-specific representation layers. In parallel, the release of T-REX (2018) marked a milestone in dataset scale and methodology, and spurred new exploration of few-shot learning capabilities. Since \textbf{2020}, RE research has expanded rapidly across models, datasets, and surveys. Encoder–decoder architectures such as T5 and BART have enable generative approaches that directly produce relational triples, saturating several benchmark datasets. Consequently, newer datasets have targeted more complex challenges, including multilinguality, fine-grained relations, and domain-specific extraction. In \textbf{2023}, we entered in the \textbf{LLM-age}, since the release of ChatGPT 3.5 which introduced a new wave of research based on large decoder-only models (LLMs). These models allow longer input contexts and instruction-based interaction without task-specific fine-tuning. However, their limitations—particularly in reasoning, factuality, auditability, and interpretability—highlight the importance of ongoing work on \textbf{neurosymbolic} methods combining Knowledge Graphs and Language models.

\section{Conclusion}

This survey has shown that progress in relation extraction (RE) is closely tied to the quality of datasets and the ability of models to manage noisy input. Over time, datasets and models have iteratively evolved to address more complex relations, domains, and tasks. In the last five years, pre-trained Transformer architectures have driven remarkable improvements, enabling a shift from classification-based designs to generative models capable of directly producing structured relational outputs. While our review could not fully capture the most recent developments in large language models (LLMs), emerging approaches—including prompt-based, instruction-tuned, and chain-of-thought methods—are rapidly reshaping the field. At the same time, there is increasing interest in integrating LLMs with knowledge graphs~\cite{10387715,10.1109/TKDE.2024.3469578,pan_et_al:TGDK.1.1.2,ma2025llmkg4qa}, with benchmarks such as LLM-KG-BENCH~\cite{Frey2023BenchmarkingAbilitiesLarge,Frey2024AssessingEvolutionLLM} tracking these capabilities. We encourage readers to consult complementary surveys on knowledge integration in LLMs~\cite{WANG2023190}, hallucination reduction~\cite{agrawal-etal-2024-knowledge}, reasoning in graph construction~\cite{10.1007/s11280-024-01297-w}, and specialised perspectives such as domain-specific RE~\cite{Wang_Yue_Duan_2023}, few-shot learning~\cite{Chen2021ZeroShotAF}, and multimodal contexts~\cite{math11081815}.

Despite these advances, several challenges remain. Many datasets, although derived from structured knowledge bases, oversimplify information: inputs are reduced to plain text and outputs are represented as triples that often fall short of Semantic Web standards. Research also remains fragmented across communities of practice. Encyclopedic datasets aim to cover entire knowledge graphs but struggle with overlapping or semantically close relations, while scientific and domain-specific corpora remain narrow in scope yet difficult for current models. Future RE systems will therefore need to remain \textbf{knowledge-grounded}, ensuring auditability and factual consistency, and increasingly \textbf{ontology-driven}, to enable principled specialisation and reuse of extracted data.

On the modelling side, transformers and LLMs have simplified task design and enabled richer relational extraction, but at significant computational cost: even adapting smaller models requires substantial GPU resources. As such, lightweight and modular architectures are valuable, particularly for constrained tasks where efficiency is paramount. In knowledge-intensive applications, RE models can be seen not only as extractors but also as structured data generators and translators within complex agent-based systems. In such contexts, frugality and scalability will be key, reinforcing the relevance of exploring small language models and hybrid approaches~\cite{belcak2025smalllanguagemodelsfuture}.

\begin{acks}
This work was supported by the French government through the France 2030 investment plan managed by the National Research Agency (ANR), as part of the Initiative of Excellence Université Côte d'Azur (ANR-15-IDEX-01). Additional support came from French Government's France 2030 investment plan (ANR-22-CPJ2-0048-01), through 3IA Côte d'Azur (ANR-23-IACL-0001).
\end{acks}

\bibliographystyle{ACM-Reference-Format}
\bibliography{biblio3}

\appendix

\end{document}